\documentclass[10pt,journal]{IEEEtran}
\usepackage{algorithm}
\usepackage{algorithmic}
\usepackage{graphicx}
\usepackage{siunitx}
\usepackage{url}
\usepackage{amsmath}
\usepackage{soul}
\usepackage{subcaption}
\usepackage{hyperref}

\usepackage{graphicx}
\usepackage{etoolbox}
\newcommand{\insertfig}{\setcounter{figure}{0}\includegraphics[width=0.95\linewidth]{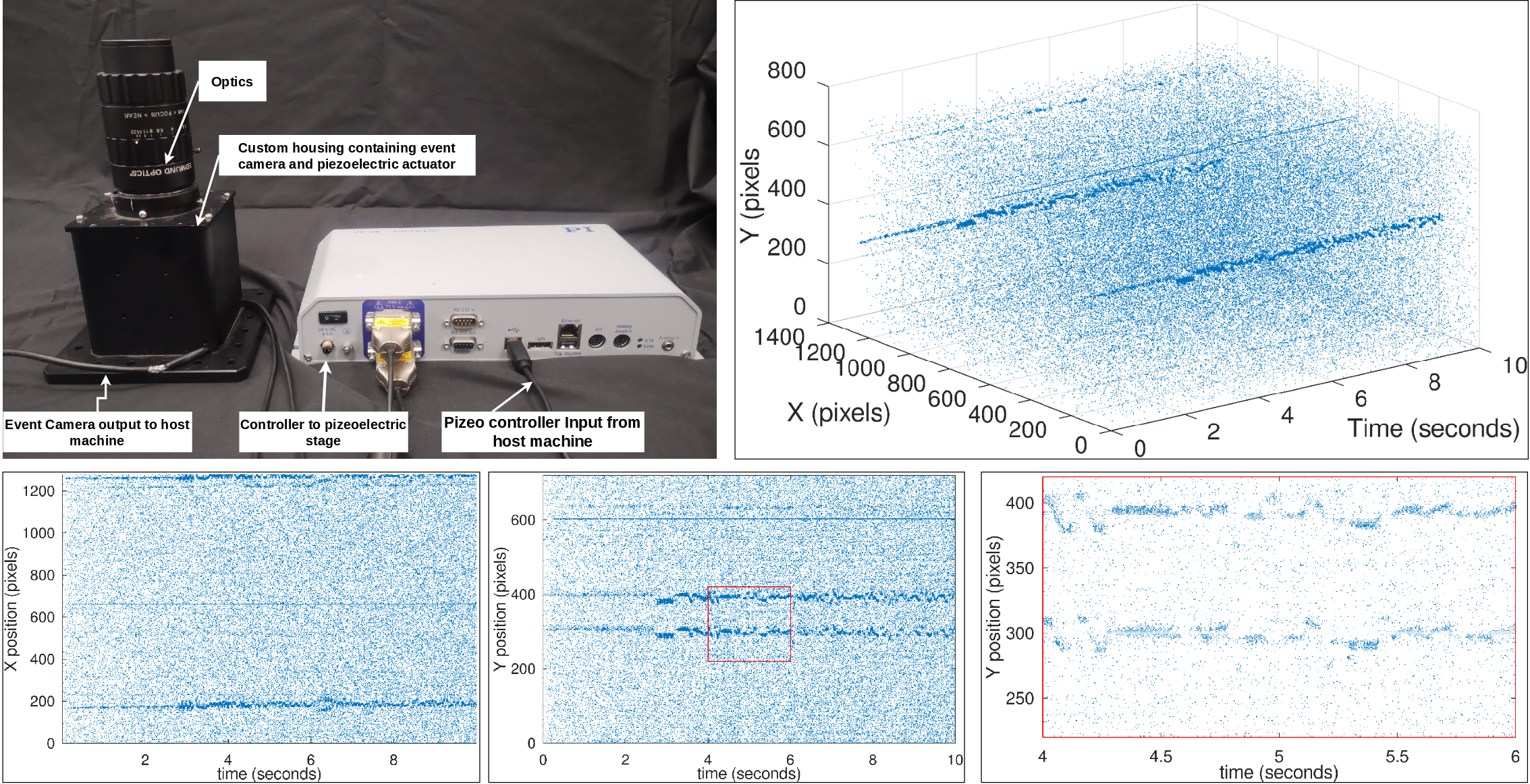}
\captionof{figure}{The e-STURT Dataset: \textbf{Top row}) \textbf{L:} An event camera mounted on a 2-DoF piezo-electric stage allows high-frequency jitter to be added to real star observations via an event camera \textbf{R:} The resulting event stream visualized as a spatiotemporal volume (XYT). Dense clusters correspond to stellar observations and introduced jitter is easily observable.
\textbf{Bottom row}: \textbf{L-R:} XT, YT view of the spatiotemporal event stream shows the magnitude of jitter along the X and Y axis respectively over time. Marked regions is shown on (\textbf{R}) illustrating the jitter for two observed stars.}\label{fig:teaser}
}% define the image

\makeatletter
\apptocmd{\@maketitle}{\centering\insertfig}{}{}% insert the figure after authors
\makeatother

\begin{document}
\title{Event-based Star Tracking under Spacecraft Jitter:  the e-STURT Dataset}

\author{
    \IEEEauthorblockN{Samya Bagchi\IEEEauthorrefmark{1}, Peter Anastasiou\IEEEauthorrefmark{2}, Matthew Tetlow\IEEEauthorrefmark{2}, Tat-Jun~Chin\IEEEauthorrefmark{1},
    Yasir Latif\IEEEauthorrefmark{1}}\\
    \IEEEauthorblockA{\IEEEauthorrefmark{1}Australian Institute for Machine Learning}
    \IEEEauthorblockA{\IEEEauthorrefmark{2}Inovor Technologies}
}

% \author{Samya~Bagchi$^1$,
%         Peter Anastasiou$^2$,
%         Matthew Telow$^2$,
%         ~Tat-Jun~Chin$^1$,
%         and Yasir~Latif$^1$% <-this % stops a space\\
%         \\$^1$Australian Institute for Machine Learning, $^2$Inovor Technologies.
% \thanks{Authors are with the Australian Institute for Machine Learning.}% <-this % stops a space
% \thanks{\texttt{firstname.lastname@adelaide.edu.au}}% <-this % stops a space



\maketitle
\begin{abstract}
Jitter degrades a spacecraft's fine-pointing ability required for optical communication, earth observation, and space domain awareness. Development of jitter estimation and compensation algorithms requires high-fidelity sensor observations representative of on-board jitter.
%These tasks demand sub-arc-second accurate pointing and stable attitude control, which are often compromised by spacecraft jitter from internal (reaction wheels) as well as external sources (atmospheric drag). Jitter degrades sensor performance reducing effective resolution and leading to pointing errors that pose significant challenges for modern space operations. Spacecrafts rely on on-board star trackers to estimate their attitude and thus pointing. CMOS-based star trackers operate at low frequencies (30 Hz), limiting their ability to measure and compensate high-frequency jitter. 
%
%Neuromorphic event sensors offer a promising alternative due to their asynchronous operation, microsecond temporal resolution, and high dynamic range. By enabling low-latency high-frequency perception,
%
%event sensors have the potential to complement traditional star trackers in precise jitter estimation and in-the-loop compensation, leading to improved pointing performance. However, event sensors are still in their infancy for space applications and event data from actual missions is not publicly available to analyze jitter characteristics and develop mitigation algorithms.
%
%
In this work, we present the \textit{Event-based Star Tracking Under Jitter} (e-STURT) dataset -- the first event camera based dataset of star observations under controlled jitter conditions. Specialized hardware employed for the dataset emulates an event-camera undergoing on-board jitter. While the event camera provides asynchronous, high temporal resolution star observations, systematic and repeatable jitter is introduced using a micrometer accurate piezoelectric actuator. Various jitter sources are simulated using distinct frequency bands and utilizing both axes of motion. Ground-truth jitter is captured in hardware from the piezoelectric actutor. The resulting dataset consists of 200 sequences and is made publicly available\footnote{\url{https://zenodo.org/records/14031911}}. 
This work highlights the dataset generation process, technical challenges and the resulting limitations. % 
% Ground truth for the introduced jitter is captured in hardware from the piezoelectric actuator. 
% 
To serve as a baseline, we propose a high-frequency jitter estimation algorithm that operates directly on the event stream. The e-STURT dataset will enable the development of jitter aware algorithms for mission critical event-based space sensing applications. 

\end{abstract}

% \begin{IEEEkeywords}
% IEEE, IEEEtran, journal, \LaTeX, paper, template.
% \end{IEEEkeywords}

% \fi
\IEEEpeerreviewmaketitle

\section{Introduction}
Human utilization of space is rapidly increasing. Modern technologies such as high-speed optical communication~\cite{Kaushal_Kaddoum_2017}, low-latency space situational awareness (SSA)~\cite{Wang_2019}, and space-based Earth observations~\cite{Tong2014FrameworkOJ} require spacecrafts to meet stringent pointing requirements~\cite{latif2023high} -- the spacecraft should be able to precisely estimate and accurately maintain its attitude (i.e., orientation) during operation. Precise maneuvering for collision avoidance~\cite{durali2006collision}, in-orbit refueling to extend a spacecraft's life~\cite{Flores-Abad_etal_2014}, and high-speed in-space optical communication~\cite{Kaushal_Kaddoum_2017}, all require sub-arcsecond pointing accuracy~\cite{Markley_Crassidis_Book_2014}. %Similarly, space-based Earth observation missions, critical for atmospheric composition mapping~\hl{[ref]}, disaster monitoring~\hl{[ref]}, and infrastructure assessment~\hl{[ref]}, rely on the spacecraft being able to maintain pointing errors below 10 microradians~\hl{[ref]}. 

%Where does jitter come from?
Orbits closer to the Earth, the Low Earth Orbits (LEO), offer low-latency communication, higher resolution imagery, and lower launch costs. However, spacecrafts in LEO travel through a thin layer of the Earth's atmosphere, leading to additional drag and vibrations in the spacecraft body~\cite{pan2017satellite, wang2021jitter}. 
%This thin layer of atmosphere acts as the largest source of external perturbation~\hl{[ref]} for spacecrafts in LEO. 
Moreover,  spacecraft components such as Cryocooler~\cite{cryocoolersVibrations}, reaction wheels~\cite{wang2021jitter}, and solar panels~\cite{solarpanelVibrations} generate vibrations in various frequency bands. While the spacecraft is extensively tested to withstand vibrations, sensors are still affected by the induced jitter, degrading the quality of observations and reducing pointing accuracy~\cite{Dennehy2019ASO, dennehy2021spacecraft}. For example, the Solar Dynamics Observatory (SDO)~\cite{pesnell2012solar}, one of the largest solar observing spacecraft ever placed into orbit, faced jitter-induced challenges affecting its high-gain antenna, responsible for transmitting more that 2 terabytes of data each day back to Earth. Jitter caused the antenna to momentarily deviate from its target direction, leading to data degradation and signal loss.

Jitter affects all sensors on-board a spacecraft, including the
Attitude Determination and Control System (ADCS) that determines the spacecraft's attitude (orientation) in space. The ADCS employs multiple sensors including star trackers, inertial measurement units (IMUs), and sun and moon sensors~\cite{Markley_Crassidis_Book_2014}, to estimate the spacecraft's orientation. While most sensors are specialized to operate under particular conditions -- the sun and moon sensors -- star trackers are exteroceptive general purpose sensors that estimates the orientation of the spacecraft from visible stars~\cite{Kim2015OPTICALJE}. Jitter affects these sensors, contributing to errors in attitude estimation. Traditional star trackers, based on CMOS/CCD-sensors~\cite{liebe2002star}, operate at relatively low frequencies (1–10 Hz)~\cite{Jun:15} to ensure higher signal to noise ratio when detecting fainter stars. This limits their ability to detect and mitigate jitter which in generally high frequency~\cite{Dennehy2019ASO}.

%Among other sensors, the ADCS contains a star tracker which estimates the .
% They operate by comparing images of stars to determine the satellite's instantaneous attitude. 
%Star trackers form part of the Attitude Determination and Control System (ADCS), enabling real-time corrections to maintain precise satellite orientation. 

%This discrepancy creates a significant blind spot in attitude determination systems, particularly for detecting and compensating high-frequency vibrations.
% Furthermore, traditional star trackers have high power consumption, typically 5–15 W during operation,~\cite{Clark2001DesignAP}, which can constitute  up to 10\% of a small satellite's power budget. 
%They are also sensitive to disturbances such as stray light, further compromising the system performance~\cite{nam2025jitterjigsawtemporaltransformer}. %Despite these limitations, 

% Overall, the star tracker is best suited for whole-body attitude control applications -- the ADCS maintains overall spacecraft orientation (e.g., for ground-station communication). However, for fine pointing tasks where a particular sensor of interest needs to be considered, sensor-specific jitter estimation~\cite{latif2023high} is required for better jitter characterization. This sensor-focused approach facilitates allows incorporation of novel sensors for jitter estimation, improves captured data quality and can provide additional inputs for whole-satellite attitude estimation.
Recently, neuromorphic event sensors have been explored for space applications. Instead of generating an image at a regular cadence, they report intensity changes as `` events''. Pixels in the event camera operate independently and asynchronously, and it offers microsecond-resolution~\cite{Benson2023} sensing. This ($\mu$s) temporal resolution enables high-frequency perception, far above the traditional CMOS/CCD star trackers, 
and enables sensor-based jitter estimation.
In addition, event cameras offer low power consumption and high dynamic range, making them particularly suitable for resource-constrained spacecraft platforms~\cite{Benson2023, cohen2017event}. %Event-based methods thus have significant potential to enhance satellite jitter estimation while reducing energy consumption and data processing requirements.
Since event sensors have not been extensively employed in active missions, the availability of space-borne data for jitter characterization and estimation is limited. This work address this challenge by presenting
%
% This work evaluates the feasibility of using event-based cameras to estimate high-frequency satellite jitter. Due to a lack of available real-world event-camera datasets from space environments, our first contribution is 
%
the \textit{Event-based Star Tracking UndeR jiTter} (e-STURT) dataset that consists of  event-based observations of real stars under controlled jitter. Specialized hardware utilized for the datasets consists of an event camera mounted on a pizeoelectric actuator. Using an event sensor for observations and the piezoelectric actuator for characteristic jitter generation places this dataset close to space-borne observations.
%
% Capturing data through the sensor, instead of simuation, enalbes
%
% The dataset consists of sequences in which the event sensor observes real stars, to capture sensor-specific disturbances such as noise -- a challenging phenomena to replicate accurately through simulation based systems. %Thus, our hardware-based approach closely replicates realistic onboard observational conditions.
%
%
% The system introduces controlled vibrations via the actuator to simulate onboard jitter and captures the resulting events using the event camera (Sec. ). 
%
% Such data enables the evaluation of event sensors for jitter mitigation. 
%Reducing the number of independent modules in space applications is critical, as it lowers the risk of component failure and increases overall system reliability~\cite{Dennehy2019ASO}.
%
Specifically, our work makes the following contributions:

\begin{enumerate}
    \item %\textbf{Jitter Dataset with Precise Ground Truth}:
   We introduce a novel hardware setup for emulating satellite jitter using an event-based camera mounted on a piezoelectric actuator (see Fig.~\ref{fig:teaser}). Accurate ground truth for jitter is captured from the piezoelectric actuator. 
   \item Using the hardware, we capture a comprehensive dataset spanning 200 sequences with controlled jitter in varying frequency bands (0–30 Hz, 30–100 Hz, and 100–200 Hz) to simulate various sources. The resulting dataset will be made public on acceptance.
    \item %\textbf{Event-Driven Jitter Recovery Algorithm}:  

   Lastly, we present an event-based algorithm for jitter estimation. Results are presented for the proposed dataset where our method demonstrates effective real-time detection and correction of jitter. 
   %without requiring additional active vibration mitigation hardware—representing a significant advancement over conventional methods in dynamic environments.
\end{enumerate}

%This work serves two main aims: provide a first dataset to study the effect of jitter on space-borne platforms and to allow development of effective jitter detection and mitigation algorithms for the event sensing and space communities. 

The remainder of this paper is organized as follows: Sec.~\ref{sec:related_works} reviews existing approaches toward star tracking and methods for jitter mitigation; Sec.~\ref{sec:dataset_generation} describes the dataset collection methodology; and  Sec.~\ref{sec:dataset} details the e-STURT dataset. Sec.~\ref{sec:algorithm} presents the proposed jitter recovery algorithm; Sec.~\ref{sec:results} presents experimental results demonstrating the algorithm's effectiveness at estimating jitter across various frequency bands.

\section{Related Work}\label{sec:related_works}

%Onboard jitter is a critical issue for spacecrafts, significantly impacting mission performance as it degrades pointing accuracy and effects the quality of imagery captured from on-board sensors~\cite{Benson2023, Kim2015OPTICALJE}. 

Various strategies have been developed in literature to address spacecraft jitter, including preventing vibration sources through improved design, mitigation via passive techniques like dampers and isolators, active estimation using additional sensors, and compensation via adaptive control systems.

\subsection{Vibration Mitigation}
Vibration mitigation -- reducing the effect of vibrations -- strategies can be broadly categorized into active and passive techniques: \textbf{Active methods} utilize external energy to counteract vibrations in real-time, offering greater control and adaptability. Common active vibration control approaches include piezoelectric actuators that convert electrical signals into mechanical deformations for precise structural vibration control~\cite{Sun2020VibrationSF}, control moment gyroscopes (CMGs)~\cite{Sun2020VibrationSF} used primarily for attitude control but also leveraged for vibration suppression~\cite{Sun2020VibrationSF}, and thrusters whose pulses can be optimally tuned using techniques such as the state-dependent Riccati equation (SDRE)~\cite{jia2015state} to suppress vibrations~\cite{XING20241}. \textbf{Passive methods}, on the other hand, do not require external energy inputs and are generally simpler to implement. Standard passive techniques include viscous damping, utilizing materials that dissipate vibrational energy through internal friction~\cite{Sun2020VibrationSF}; particle impact damping (PID), which absorbs vibrational energy through particle impacts~\cite{van2016development} within a cavity; and optimized structural design involving careful selection of materials, geometry, and mounting points to minimize vibration transmission~\cite{XING20241}. However, both passive and active vibration mitigation methods have inherent limitations. Passive methods reduce vibrations without external power but lack adaptability, often insufficient for mitigating high-frequency or multi-axis vibrations. Conversely, active methods offer real-time control but introduce additional complexity regarding power consumption, system integration, and reliability concerns due to increased hardware complexity~\cite{Xu2019}.

\subsection{Vibration Isolation}
While vibration mitigation applies broadly to spacecraft bodies, vibration isolation systems specifically separate sensitive payloads from vibration sources such as reaction wheels~\cite{belvin1995spacecraft}, control moment gyroscopes~\cite{gill2022review}, and cryocoolers~\cite{Jedrich2002CryoCI}. These systems typically use mechanical dampers or specialized isolation mounts to reduce transmission of vibrations from sources to sensitive instruments. For instance, Moog's SoftRide isolators~\cite{McVittie2013DevelopmentAP} have been successfully employed in missions like the Hubble Space Telescope~\cite{Jedrich2002CryoCI} to mitigate jitter by isolating solar arrays from the main telescope body, enhancing attitude stability and imaging performance.

\subsection{Vibration Estimation}
When direct mitigation or isolation is insufficient or infeasible, onboard sensors must estimate residual vibrations for subsequent correction or compensation.
Image-based post-processing techniques are commonly employed for jitter detection and compensation. Image matching techniques rely on multi-spectral or stereo images with rational polynomial coefficients (RPCs)~\cite{shen2017correcting} to align images accurately and detect jitter-induced distortions. Phase correlation matching has demonstrated improved accuracy in obtaining jitter curves, significantly reducing amplitude errors~\cite{isprs-annals-V-3-2022-611-2022} compared to traditional methods.
Similarly, parallax image analysis leverages displacement between adjacent images captured at different times or by different sensors to estimate jitter characteristics when high-accuracy attitude sensors are unavailable ~\cite{Pan2017SatelliteJE}. Parallax maps derived from multispectral satellite images have been successfully used to estimate jitter frequency and amplitude with high precision~\cite{rs11010016}.

\subsection{Event-Based Cameras in Space}
Event-based cameras are increasingly recognized for their potential in space applications due to several distinct advantages compared to traditional frame-based sensors. They generate asynchronous streams of events triggered by brightness changes at pixel-level resolution, achieving sub-millisecond temporal precision suitable for capturing rapid visual scene changes~\cite{Roffe_2021}\cite{Wang_2019}. Additionally, event cameras offer a high dynamic range exceeding 120 dB, enabling effective operation under extreme lighting variations common in space environments~\cite{Wang_2019}~\cite{Gallego_etal_2022}. Their sparse output significantly reduces data transmission requirements and power consumption—critical advantages for resource-constrained space missions~\cite{Gallego_etal_2022}. Furthermore, event-based cameras exhibit robustness against motion blur—essential for applications involving fast-moving objects such as spacecraft rendezvous maneuvers or asteroid tracking~\cite{Gallego_etal_2022}. However, these sensors also face limitations including sensitivity to noise under low-light conditions and susceptibility to radiation-induced effects in space~\cite{xiao2022preliminaryresearchspacesituational}.
\begin{table}[t]
    \centering

    \begin{tabular}{l|c}
    \hline
Resolution & 1280 x 720 pixels \\\hline
Physical Chip size & 6.22 x 3.5 mm$^2$ \\\hline
Pixel size & 4.86 x 4.86 um$^2$ \\\hline
Max. Bandwidth & 1066 Meps over USB3 \\\hline
Angular FOV (100 mm optics) & 3.42 degrees x 1.89 degrees \\\hline
    \end{tabular}
        \caption{Specifications of the Prophesee Gen4 HD event camera}
    \label{tab:event_camera_specs}
\end{table}

Event-based cameras have shown promise across various space applications. They have been utilized for asteroid detection and tracking from space-based platforms~\cite{Roffe_2021} due to their low latency and high temporal resolution capabilities. In spacecraft attitude determination systems, event cameras have demonstrated potential for star tracking tasks due to their robustness against motion blur and rapid response times under challenging illumination conditions~\cite{Chin_etal_2020_EventST}. Additionally, they have been applied successfully in visual odometry algorithms designed specifically for resource-limited spacecraft platforms operating under challenging lighting scenarios~\cite{8100099}. Event-based cameras' high temporal resolution also makes them attractive candidates for precise relative pose estimation during spacecraft rendezvous operations—critical maneuvers requiring rapid sensor response times with minimal latency~\cite{Wang_2019}\cite{inproceedingsMohsi2023}. Furthermore, neuromorphic vision sensors based on event-driven technology have been explored extensively for planetary exploration missions where terrain classification through spatiotemporal event patterns helps identify navigational hazards efficiently in real-time scenarios~\cite{Azzalini2023GeneratingAS}.

Previous work has also explored event cameras specifically for fine-pointing problems~\cite{latif2023high} onboard small satellites (nanosats), where instantaneous corrections required by payload sensors are estimated using event-driven sensing combined with piezoelectric actuation systems. The work in~\cite{latif2023high} is closely related to our application setting; however, their approach operates primarily on simulated star data observed via an event sensor rather than actual star observations. In contrast, our work generates real-world event streams from actual star fields observed under controlled jitter conditions using piezoelectric actuators explicitly designed for introducing precise vibrations rather than applying corrective attitude adjustments as is the objective of~\cite{latif2023high}.

\section{Dataset Generation: Hardware and Methodology}\label{sec:dataset_generation}
% This section discusses the hardware setup, various calibrations, and collection protocols for the e-STURT dataset.
\subsection{Hardware setup}
\label{sec:hardware_setup}
    The hardware setup consists of a Prophesee Gen4 HD event camera (Table.~\ref{tab:event_camera_specs}) mounted on a 2 translational degree of freedom (DoF) piezoelectric motion stage, the U-723 PILine® XY Stage from Physik Instrumente (PI) using the C-867.2U2 PILine® Motion Controller (Fig.~\ref{fig:teaser}). The optics consist of a 100mm/F2.8 lens from Edmond optics, resulting in an effective FoV of 3.42 x 1.89 degrees for the event sensor. This FoV allows each star to span at least 2 pixels in the event camera. Specifications for the event sensor and the piezoelectric stage are given in Tables.~\ref{tab:event_camera_specs} and~\ref{tab:piezo_specs}, respectively.  
%
% \subsubsection{Event Camera}
The event sensor is rigidly mounted onto the piezoelectric stage and is housed inside a weather and lightproof enclosure. 
%Light from stars focuses on the event sensor using a narrow field of view lens with an effect angular FoV of 2 degrees by 3 degrees. 
%
\begin{table}
    \centering

    \begin{tabular}{l|c}
    \hline
        Active axes & x, y \\\hline
Motion range & 22 mm × 22 mm\\\hline
Velocity, closed loop & 200 mm/s (max ) \\\hline
Bidirectional repeatability & ±0.2 um \\\hline
Load capacity in z (Payload capacity) & 10 N  (max) \\\hline
Minimum incremental motion (motion
resolution) & 0.1 um \\\hline
    \end{tabular}
    \caption{Specification of the  U-723 PILine® XY Stage from Physik
Instrumente (PI).}
    \label{tab:piezo_specs}
\end{table}
\subsubsection{Calibration and Environmental Conditions}
\label{sec:calibration}
% \begin{enumerate}
%     \item \textbf{Motion-to-Pixel Calibration}:
The first step in the calibration process is to map the displacement of the actuator to the corresponding displacement in the event sensor frame.
The piezoelectric stage’s mechanical displacement was mapped to pixel shifts using a checkerboard pattern under uniform motion. For each axis, the stage was displaced in $10~\mu$m increments over its 22~mm range while recording events. Linear regression confirmed a displacement-to-pixel ratio of $0.1$~mm~$\approx 20.58$~pixels~($R^2 = 0.998$), consistent with the sensor’s pixel size ($4.86~\mu$m). 

A spirit level was used to ensure the piezoelectric stage remained parallel to Earth’s surface during data collection (Nov 2023–Jan 2024, Adelaide, Australia). This configuration simulates orbital jitter without gravitational coupling effects absent in space environments. To minimize variations in sensor noise, data was collected under clear skies (temperature: $15^\circ\text{C}$–$25^\circ\text{C}$, humidity: $30\%$–$50\%$) with minimal light pollution. Lightproof enclosure mitigated stray light (Fig.~\ref{fig:teaser}).

\subsection{Sensor Bias Optimization for Star Detection}
\label{sec:sensor_bias_optimization}

Biases refer to configurable electrical parameters that control the sensor's sensitivity and operational characteristics, directly influencing how the event camera detects and responds to changes in scene illumination~\cite{mcreynolds2023demystifying}. They dictate trade-offs between noise suppression, event generation rates, and detection thresholds. It is vital to tune these parameters to our observation scenario so that enough events are generated for reliable star tracking under varying jitter.

The Prophesee Gen4 HD event camera's performance was optimized for star detection by systematically adjusting its bias parameters.  We employed a grid search, guided by the manufacturer's specifications and prior research on event-based star tracking~\cite{mcreynolds2023demystifying}.  The optimal bias settings, presented in Table.~\ref{tab:bias_settings}, were selected to maximize visual contrast and the number of visible stars under both static (no jitter) and dynamic (jitter) conditions. The ranges and step sizes for theses parameters are detailed below:

\begin{table}[h!]
\centering

\begin{tabular}{l|l|c}
\hline
Parameter        & Range     & Step Size \\ \hline
\texttt{bias\_diff\_on/off} & 50 - 150  & 5         \\ \hline
\texttt{bias\_refr}      & 5 - 30    & 5         \\ \hline
\texttt{bias\_pr}        & 100 - 150 & 10        \\ \hline
\end{tabular}
\caption{Grid search parameter ranges and step sizes.}
\label{tab:grid_search_params}
\end{table}

Bias configurations were evaluated based on two metrics: \textbf{a)} star count: the mean number of  stars detected per 3-minute sequence under no-jitter (static) conditions and \textbf{b)} Signal-to-Noise Ratio (SNR): the ratio of events originating from the stars against those from noise during jitter scenarios. This quantifies the camera's ability to distinguish star against the background noise under dynamic conditions.

% \subsubsection{Frequency-Specific Tuning}
% \label{subsec:frequency_tuning}

The bias settings were further fine-tuned for each of the three frequency bands (Sec.~\ref{sec:frequency_band_selection}) to balance star detection sensitivity and motion blur.
At low frequency (0-30 Hz), higher contrast thresholds (\texttt{bias\_diff\_on} = 96 and \texttt{bias\_diff\_off} = 60) improved sensitivity to faint stars, maximizing the number of detectable stars in this relatively slow-motion regime. However, at higher frequencies (100--200 Hz), lower refractory periods (\texttt{bias\_refr} = 15) were crucial for minimizing motion blur during rapid jitter, ensuring that individual star events were distinguishable. 

The grid search and subsequent frequency-specific tuning yielded the following key results: for the no-jitter case,
refined biases increased the mean star count by approximately 25\% compared to the default camera settings (from an average of 18.7 stars to 23.4 stars), and for the higher frequency jitter (100 -- 200 Hz), the SNR improved by approximately 40\% due to adjustments in \texttt{bias\_refr}.

\noindent
\textbf{Contrast Sensitivity Optimization:} The contrast sensitivity threshold biases (\texttt{bias\_diff\_on} and \texttt{bias\_diff\_off}) were set to relatively high values (96 and 60, respectively) as determined by the grid search and frequency specific tuning, to enhance the sensor's ability to detect subtle changes in illumination~\cite{prophesee_biases}. This configuration improves the camera's ability to distinguish faint stars from the background, as even minor variations in brightness will trigger events. Typically, between 5-10 stars were visible within the camera's FOV, depending on the specific region of the sky being observed.

\noindent
\textbf{Bandwidth and Noise:} The bandwidth biases (\texttt{bias\_fo\_n} and \texttt{bias\_hpf}) were both set to 0 ~\cite{prophesee_biases}, maximizing the sensor's bandwidth. This allows the sensor to detect a wider range of illumination changes, including both rapid and slow variations, which is essential for detecting stars with varying brightness and under jitter conditions.

\noindent
\textbf{Event Rate:} The refractory period (\texttt{bias\_refr}) was set to a relatively low value of 15 ~\cite{prophesee_biases}, as determined by the grid search. This reduces the duration for which a pixel is ``blind'' (inactive) after generating an event, enabling more frequent event generation. For star detection, this enables the sensor to capture more events from faint stars, particularly under jitter, where the star moves rapidly across the pixel.

It should be pointed out that at higher jitter frequencies, the number of detectable stars decreases. For rapid motion, individual pixels do not receive enough photons to trigger events. The chosen biases (Table.~\ref{tab:bias_settings}), therefore, represent a compromise, aiming to maximize star detection capabilities across the various frequency ranges.

\begin{table}[t]
\centering
%\resizebox{\textwidth}{!}{

\begin{tabular}{l|l|c}
\hline
Bias Parameters & Adjusts & Value \\ \hline
\texttt{bias\_diff} & the contrast threshold & 69 \\ \hline
\texttt{bias\_diff\_off} & contrast threshold for OFF events & 60 \\ \hline
\texttt{bias\_diff\_on} & contrast threshold for ON events & 96 \\ \hline
\texttt{bias\_fo\_n} & the low-pass filter & 0 \\ \hline
\texttt{bias\_hpf} & the high-pass filter & 0 \\ \hline
\texttt{bias\_pr} & photoreceptor bandwidth & 131 \\ \hline
\texttt{bias\_refr} & adjusts the refractory period & 15\\ \hline
\end{tabular}
\caption{Bias Settings for the dataset}
\label{tab:bias_settings}
%}

\end{table}

\subsection{Jitter Frequency Band Selection}
\label{sec:frequency_band_selection}
The selection of frequency ranges is a critical towards the impact, relevance, and applicability of e-STURT. Guided by existing literature~\cite{Dennehy2018SpacecraftMA, rs14020342}, we consider three bands: low (0--30 Hz), medium (30 -- 100 Hz), and  high (100--200 Hz) :

\subsubsection{Low-Frequency Band (0-30 Hz)}
The 0-30 Hz range captures low-frequency disturbances that are particularly relevant to satellite operations~\cite{Dennehy2019ASO, Gordon2015BenefitsOS}. This band typically includes the fundamental structural modes of the spacecraft, which can be excited by various onboard activities~\cite{Dennehy2019ASO}. The bandwidth of most satellite attitude control systems falls within this range, making it crucial for understanding control-structure interactions~\cite{Gordon2015BenefitsOS}. Additionally, solar array drive mechanisms often operate at frequencies below 10 Hz and can contribute significantly to low-frequency jitter.~\cite{Dennehy2019ASO, Gordon2015BenefitsOS}

\subsubsection{Mid-Frequency Band (30-100 Hz)}: The 30-100 Hz range encompasses a variety of important jitter sources. This is the primary frequency range for reaction wheel and cryocoolers disturbances, making this band crucial for studying one of the most significant sources of satellite jitter~\cite{kamesh2012passive}. Liquid propellant movement can also generate disturbances in this frequency range, especially during maneuvers~\cite{Hasha2016HighPerformanceRW}.

\subsubsection{High-Frequency Band (100-200 Hz)}: The 100-200 Hz range captures high-frequency disturbances that can be particularly challenging to mitigate. This band includes many of the micro-vibrations generated by onboard equipment, which can significantly impact high-precision pointing requirements~\cite{Hasha2016HighPerformanceRW}. Higher harmonics of lower-frequency disturbances also often fall within this range, potentially causing resonance with structural components~\cite{wang2021jitter, Hasha2016HighPerformanceRW}.
When present, Control moment gyroscopes (CMGs) can introduce high-frequency disturbances also within this range~\cite{vibration7010013}.

% Methodological Considerations:
% The selection of specific frequency bands was driven by several factors.
% %\begin{itemize}
% %\item Comprehensive coverage: 
% The 0 to 200 Hz band ensures coverage of the most significant jitter sources identified in spacecraft literature~\cite{Pan2017SatelliteJE, ijgi13110413}.
%\item Alignment with sensor capabilities: 
% The chosen frequency bands correspond to actual disturbance sources encountered in satellite operations,  such as the Hubble Space Telescope and the Solar Dynamics Observatory~\cite{Dennehy2019ASO, liu2008reaction}.
% Event-based cameras have demonstrated effectiveness in capturing high-frequency vibrations up to 190 Hz~\cite{ Pan2017SatelliteJE}, making our chosen ranges well-suited to the sensor's capabilities.
%\item Data quality optimization: 
% Validation tests indicate that these ranges produced the most consistent and analyzable event data, balancing event generation and motion blur avoidance~\cite{Pan2017SatelliteJE}.
%\item Relevance to real-world scenarios: 
%\end{itemize}
% By focusing on these specific frequency bands, our research aims to provide a comprehensive understanding of satellite jitter across the most relevant spectral ranges. It enables the development and validation of jitter recovery algorithms that can address the full range of vibration challenges encountered in space missions.

\begin{figure}
    \centering
    \fbox{
    \includegraphics[width=0.8\linewidth]{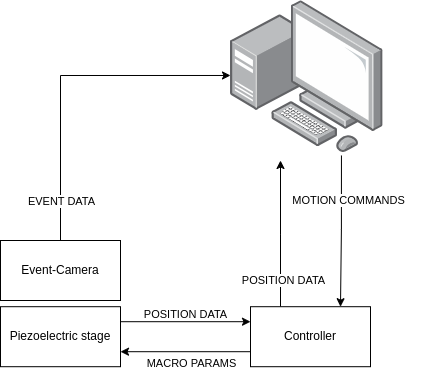}}
    \caption{Flow of information between the host (computer), controller and  the piezoelectric stage. The host communicates with the controller to provide motion commands and macro parameters. Macros reside on and are executed on the piezoelectric stage, allowing it to experience controlled jitter. This leads to an event stream that is recorded by the host machine.}
    \label{fig:dataflow}   
\end{figure}

\subsection{Velocity and amplitude selection}
% Our experimental setup and methodology were carefully designed based on both preliminary tests and insights from recent literature on event cameras and vibration measurement.
 Event camera triggers event on intensity change at a pixel. This happens when either the scene changes or the event-sensor undergoes egomotion, as is the case in our setting. With default biases, movements smaller than $0.1$ mm did not consistently produce  sufficient events. Default biases are tuned to  filter noise and minor fluctuations. However, carefully selecting biases (Sec.~\ref{sec:sensor_bias_optimization}) ensures that $0.1$ mm movements consistently generates enough events to accurately characterize jitter. Movement of $0.1$ mm corresponds to approximately 20.58 pixels in the event-sensor frame (Sec.~\ref{sec:calibration}). 
 %The choice of $0.1$ mm aligns with previous findings~\hl{[ref]}. 
 We term this the motion amplitude $a$.
%Our approach aims to address this by carefully calibrating movement speed to ensure consistent event generation.
%
Given a fixed amplitude $a$, the velocity $v_f$ at a frequency $f$ is can be computed as 
\begin{equation}
    v_f = f/(2*a)
\end{equation}
\noindent
This determines the minimum and maximum velocity for each frequency band reported in Table.~\ref{tab:settings_velocity}.

\begin{table}[b]
    \centering

    \begin{tabular}{c|c|c}
        Setting & Velocity Range (mm/s) & Frequency Range (Hz)\\\hline
        \texttt{slow} & $1$--$6$ &  $0$-$30$ \\\hline
        \texttt{medium} & $6$--$20$  & $30$-$100$ \\\hline
        \texttt{fast} & $20$--$40$  & $100$-$200$ \\    
    \end{tabular}
    \caption{Velocity and frequency parameters}
    \label{tab:settings_velocity}
\end{table}

\subsection{Jitter Generation}
\label{sec:jitter_generation}
Precise control of the piezoelectric vibrator is crucial for accurately emulating satellite jitter. 
Under normal operation, the actuator receives motions commands from the host machine via controller, executes them and reports back its position. The controller acts as an intermediary between the actuator and the host machine.
However,
high-frequency jitter generation presents  challenges in such a simple setup due to inherent communication delays between the host computer, controller, and piezo vibrator (Fig.~\ref{fig:dataflow}). The communication is much slower than the required response time for high-frequency motion commands to be issued, executed, and the final acknowledgment returned back to the host computer. To overcome this delay, we developed a novel approach to low-latency jitter generation by leveraging the on-device capabilities of the piezoelectric stage. We utilize embedded macros paired with a novel circular queue data structure. 

The controller integrated within the piezoelectric stage can store up to 350 assembly-like instructions as ``macros''. These macros can be preloaded onto the controller and executed via function calls from the host computer, significantly reducing communication delays. Macros can write output data (such as position and velocity of the actuator) to limited on-device storage consisting of only nine float variables. These memory locations can then be sequentially read by the host computer. Coupling the on-device storage with interleaved read operations allows accommodating up to 8 times slower lossless communication. However, if the write operations are much faster than that, some data loss will be inevitable.

A SET operation instructs the piezo to move to a specific (x, y) position, while a GET operation queries the current position of the piezo. To analyze the system's Input-Output (I/O) capabilities and positioning delays, we executed 10,000 iterations of SET and GET operations. The mean completion time for SET operations was approximately $15$ ms, and for GET operations, it was approximately $32$ ms. The mean and maximum times to READ (retrieving data from the registers) were $32$ ms and $40$ ms, respectively. 

\begin{figure}
    \centering
    \includegraphics[width=\linewidth, height=8em]{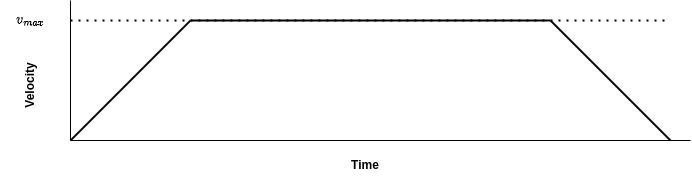}
    \caption{The velocity profile of the piezoelectric stage. From rest, the stage first accelerates to reach a predefined $v_{max}$, moves then with constant velocity, and decelerates to reach the target position.}
    \label{fig:velocity-profile}
\end{figure}
In addition to the I/O delays, we investigated the positioning accuracy and response time of the piezoelectric vibrator. The piezo stage follows a trapezoidal velocity profile (Fig.~\ref{fig:velocity-profile}), where it first accelerates to a predefined maximum velocity $v_{max}$ from rest, moves with ${v_{max}}$, and finally decelerates to reach zero velocity at the target. During the deceleration period, the piezo stage makes fine adjustments to align itself as closely as possible with the specified target position. Due to this, the piezo exhibited a significant delay in reaching the target position with a mean fine-tuning time of $25$ ms, regardless of the commanded velocity or amplitude. This positioning delay is further compounded by the direct proportionality between the velocity and positional errors, as well as the distinct error characteristics of each axis. 
Based on this comprehensive understanding of the system's delays and limitations, we developed a set of embedded macros that optimize both vibration control and data acquisition. The core vibration control logic is encapsulated in the (Algorithm.~\ref{alg:vibrate}).

\begin{algorithm}
\caption{MACRO\_VIBRATE ($amplitude$, $delay$)}
\begin{algorithmic}[1]
\STATE Move {$axis1$ by $amplitude$}
\STATE Move {$axis2$ by $amplitude$}
\STATE Delay {$delay$}
\STATE $output1 \gets AXIS1\_POS$
\STATE $output3 \gets AXIS2\_POS$
\STATE Move {$axis1$ by -$amplitude$}
\STATE Move {$axis2$ by -$amplitude$}
\STATE Delay {$delay$}
\STATE $output5 \gets AXIS1\_POS$
\STATE $output7 \gets AXIS2\_POS$
\end{algorithmic}
\label{alg:vibrate}
\end{algorithm}

The MACRO\_VIBRATE macro, designed to generate a single back-and-forth motion, is presented as an example. It instructs the piezo to move to a specified position (\texttt{amplitude}) and records the position. The \texttt{delay} allows sufficient time for the piezo to execute the motion. It then issues another set of motion commands along both axes that bring the piezo back to its original position and records the position after an additional \texttt{delay}. While such a simple macro can generate vibrations, the rate at which the positions are recorded can not be reliably read by the host due to the communication bottleneck and the asynchronous nature of macro execution. 

\begin{figure}
    \centering
    \includegraphics[width=0.75\linewidth]{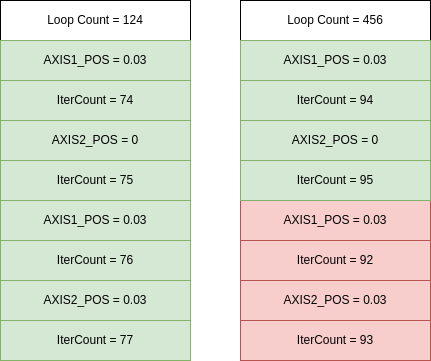}
    \caption{Sample executions of Algorithm.~\ref{alg:circular_queue}. Valid values are marked in green and are determined by ensuring that the values of \texttt{StepCount} are  in increasing order from top to bottom.}
    \label{fig:queue}
\end{figure}

To address the limitations of the basic VIBRATE macro, we designed a more sophisticated macro, MACRO\_CIRCULAR\_QUEUE (Algorithm.~\ref{alg:circular_queue}), that embeds not only position data but also information about the relative arrival order of the data in a circular buffer. This macro accepts two inputs: \texttt{amplitude} (the magnitude of motion along either axis) and \texttt{delay} (the wait time between motion executions). During macro execution, information about the axes' positions is interleaved with two counters: \texttt{LoopCount} (the overall count of macro executions) and \texttt{StepCount} (the number of motion command executions, serving as a pseudo-timestamp). Because the controller provides only nine output registers, the circular queue can store up to four axis positions and five counters (see Fig.~\ref{fig:queue}).

The validity of the data contained in the queue is determined by comparing the values of \texttt{StepCount} from top to bottom. If all values are monotonically increasing, the data in the queue is in the correct order and can be read out. Alternatively, only recent data, as indicated by consistently increasing \texttt{StepCount} values, is considered valid (as demonstrated in Fig.~\ref{fig:queue})
By associating each pair of \texttt{LoopCount} and \texttt{StepCount} with a timestamp (derived from the \texttt{delay} parameter and adjusted for I/O delays), we can reconstruct the temporal evolution of the piezo's position. This macro iterates $N$ times as dictated by the host computer. During execution, the host computer can asynchronously update additional, randomly generated velocity parameters (not shown here), with the update rate dependent on the frequency band of the current experiment (Table.~\ref{tab:settings_velocity}). This approach effectively simulates real-world satellite jitter scenarios, where vibration characteristics can change dynamically. The circular queue ensures that the host computer reads the most recent data without interrupting the ongoing vibration process, facilitating real-time data acquisition. Key parameters summarizing the e-STURT dataset are provided in Table \ref{tab:params}.

\begin{table}[t]
    \centering
    \begin{tabular}{l|l}
    \hline
        \textbf{Parameter} & \textbf{Value} \\ \hline
        Total sequences & 200 \\ \hline
        Temporal resolution & 1 $\mu$s \\ \hline
        Spatial resolution & 1280 $\times$ 720 px \\ \hline
        Jitter amplitude & 0.1 mm (20.58 px) \\ \hline
        Ground truth rate & 30 Hz \\ \hline
    \end{tabular}
    \caption{Key Dataset Parameters}
    \label{tab:params}
\end{table}

\section{The e-STURT Dataset: Design and Characteristics}\label{sec:dataset}

The e-STRUT dataset was collected over 20 nights under clear skies in Adelaide, Australia. The data acquisition was structured into episodes, each encompassing a comprehensive set of motion configurations and jitter frequency bands. To establish a baseline and characterize environmental noise, each episode started with a no jitter \texttt{static} sequence. Data was then captured for motion along the first axis (\texttt{Axis1}), the second axis (\texttt{Axis2}), and both axes simultaneously (\texttt{BothAxis}), under three distinct jitter frequency regimes: \texttt{slow} (0-30 Hz), \texttt{medium} (30-100 Hz), and \texttt{fast} (100-200 Hz). For each frequency band, motion along each axis configuration was executed sequentially. This systematic approach resulted in 10 distinct sequences per episode. Across 20 episodes, the e-STURT dataset comprises a total of around 200 sequences, providing a rich and diverse resource for jitter estimation.

\subsection{Data Acquisition Protocol}

The total duration of each sequence was set to 190 seconds. This duration was chosen to accommodate hardware initialization overhead and provide a sufficient observation window. Specifically, 180 seconds were allocated for data acquisition under controlled conditions, preceded by 10 seconds for actuator homing and stabilization, and initial camera baseline recording as described below:
%This extended capture window ensures the comprehensive recording of both the pre-jitter baseline phase (5 seconds) and the subsequent dynamic vibration phases (180 seconds).

% The data acquisition process for each sequence followed these steps:

\begin{itemize}
    \item \textbf{Event Camera Initialization}: Prior to inducing any controlled motion, each recording commenced with a 5-second capture of baseline star motion. This segment aimed to record ambient atmospheric perturbations and establish a reference for subsequent jitter analysis.
    \item \textbf{Piezoelectric Homing Sequence}: To ensure consistent and repeatable jitter induction, a homing sequence was implemented for the piezoelectric actuator. First, using the using the Physik Instrumente \texttt{FRF} command, the piezoelectric stage were driven to a reference position (0,0). Then the stabilization of both axes at the zero position was verified via \texttt{qFRF} status polling command. Once confirmed, a 5-second pause was introduced to allow the event camera to stabilize for data acquisition.
    
    \item \textbf{Primary Jitter Injection}: To initiate the controlled jitter, the following steps were executed. First, a series of ten sequential motion command of 0.1mm were issued  at maximum velocity, followed by a single synchronization spike of 0.4 mm. This spike serves as a temporal marker to precisely align the timestamps from the event camera ($t_{cam}$) and the piezoelectric actuator ($t_{piezo}$), ensuring temporal synchronization.
    
    \item \textbf{Iterative Jitter Profile}: Random jitter was then introduced by alternating commands of 0.1 and -0.1 in the axis of motion. Random velocity changes were applied to modify the jitter profile. The velocity and modulation parameters were logged in hardware using the proposed macros.
    
    % \begin{enumerate}
    %     \item \textit{Continuous Bidirectional Motion}:  Alternating \texttt{MOV 0.1mm} and \texttt{MOV -0.1mm} commands were continuously issued to the piezoelectric actuator, inducing bidirectional motion.
    %     \item \textit{Dynamic Velocity Modulation}:  Random velocity variations, within experiment-specific ranges (defined in Algorithm 1), were dynamically applied to modulate the jitter profile, simulating more realistic and complex vibration patterns.
    %     \item \textit{Real-time Parameter Updates}:  Velocity adjustments and modulation parameters were updated in real-time via embedded macros on the piezoelectric controller. This allowed for dynamic control of the jitter profile without interrupting the data acquisition process, ensuring continuous data streams reflecting evolving vibration characteristics.
    % \end{enumerate}
\end{itemize}

% \subsection{Dataset Characteristics and Sample Visualizations}

% \subsection{Dataset Organization by Jitter Configuration}
The resulting dataset exhibits diverse characteristics across different jitter frequencies and motion axes. Furthermore, each frequency band contains three axial configurations: \texttt{Axis1}where the motion happens along the x-axis of the  actuator (horizontal in the sensor plane), \texttt{Axis2} y-axis excitation creating vertical displacement in the event-sensor and finally \texttt{BothAxes} where X-Y actuation happens simultaneous leading to more complex jitter patterns. Axial and frequency bands leads to 9 different combinations of jitter generation.

% \subsubsection{Frequency-Specific Structure}
% \begin{itemize}
%     \item \textbf{Low-Frequency Band}: Contains 0-30Hz sequences for X, Y, and XY configurations.
%     \item \textbf{Mid-Frequency Band}: Contains 30-100Hz sequences for X, Y, and XY configurations.
%     \item \textbf{High-Frequency Band}: Contains 100-200Hz sequences for X, Y, and XY configurations.
% \end{itemize}

% \subsection{Static Sequence (No Jitter)}
% \begin{itemize}
% For each capture episode, the \texttt{static} sequence comprises baseline event camera data. This data captures the natural motion of stars due to sidereal movement and atmospheric perturbations, providing a reference for evaluating noise characteristics and natural star field dynamics in the absence of controlled jitter.

%\subsection{Dataset Organization and Access}

The dataset is structured into 20 experimental episodes, each containing 10 sequences categorized by frequency range and motion axes, as summarized in Table \ref{tab:bands}.

\begin{table}[t]
    \centering

    \begin{tabular}{c|c|c}
        \hline
        \textbf{Frequency range} & \textbf{Axial configurations} & \textbf{Sequences per episode} \\
        \hline
        Static Reference & None & 1 \\
        0–30 Hz & X, Y, XY & 3 \\
        30–100 Hz & X, Y, XY & 3 \\
        100–200 Hz & X, Y, XY & 3 \\
        \hline
    \end{tabular}
        \caption{Frequency Band Configuration}
    \label{tab:bands}
\end{table}

Each sequence is provided in three file formats:
\begin{itemize}
    \item \textit{Raw Event Streams (\texttt{.dat})}:  Raw event data recorded at the sensor's native resolution of $1280 \times 720$ pixels, preserving the complete asynchronous event stream.
    \item \textit{Piezoelectric Telemetry (\texttt{.csv})}:  Telemetry data from the piezoelectric actuator, sampled at 30 Hz, providing ground truth measurements of the induced jitter motion.
    \item \textit{Hardware-Synchronized Timestamps (\texttt{.log})}: Hardware-synchronized timestamps, logged to ensure accurate temporal alignment between event camera data and piezoelectric telemetry, critical for precise jitter analysis and compensation algorithm development.
\end{itemize}

\subsection{Data Anomalies and Missing Data}

It is important to note that the e-STURT dataset includes some instances of incomplete data or anomalies:

Episode 2 exhibits incomplete data for the mid and high-frequency jitter regimes due to piezoelectric actuator stiction. Specifically, only one sequence, corresponding to mid-frequency jitter along Axis 1, is available. Data for Axis 2 and Both Axes configurations are missing. Additionally, no data is available for the high-frequency jitter regime (100-200 Hz) due to actuator malfunction. 
Episode 20 deviates from the standard synchronization spike protocol implemented in later episodes. This sequence originates from an earlier experimental run and does not incorporate the synchronization spike logic introduced in subsequent data collection efforts. As a result, precise temporal alignment with ground truth data may require alternative synchronization methods.

% The stratified and comprehensive design of the e-STURT dataset facilitates systematic evaluation of vibration compensation algorithms across 27 unique jitter profiles. The inclusion of hardware-synchronized timestamps and ground truth telemetry, validated through spike-induced event clusters ($\Delta t < 33.3$ ms), enables rigorous quantitative performance assessment and algorithm benchmarking. The dataset is intended to be publicly released to foster further research and development in event-based star tracking and jitter mitigation techniques for space-borne applications.

% To ensure accurate synchronization of the axis positions with their corresponding loop counts, we implemented a circular queue using nine output variables as shown in. By carefully interleaving motion commands with data storage operations, the macro minimizes the impact of I/O delays on overall system performance.}

\begin{figure*}
    \centering
    \includegraphics[width=0.32\textwidth]{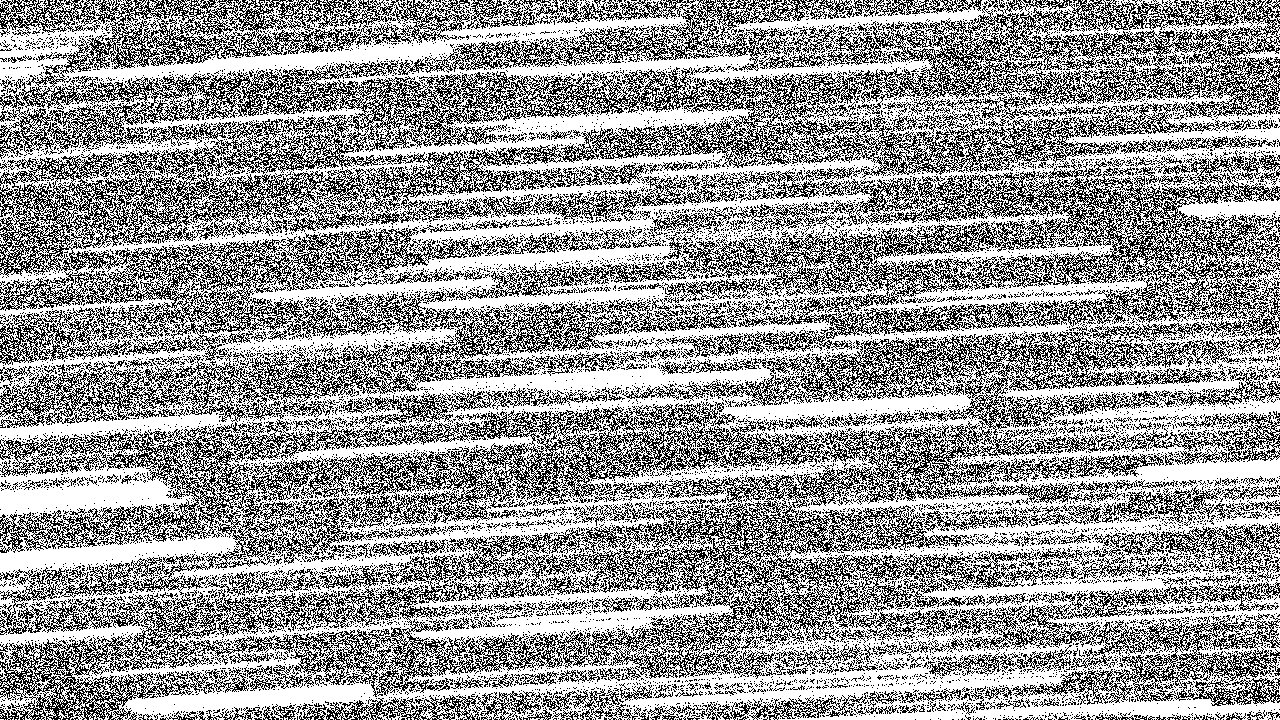}
    \includegraphics[width=0.32\textwidth]{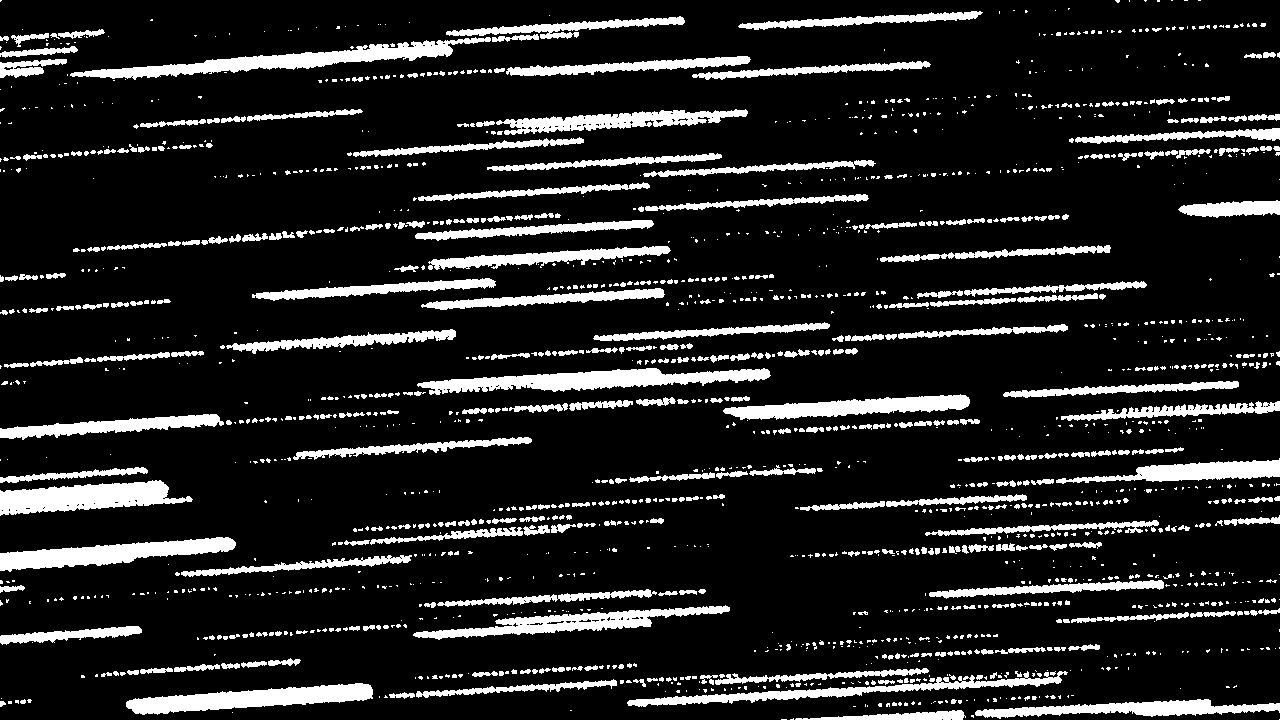}
    \includegraphics[width=0.32\textwidth]{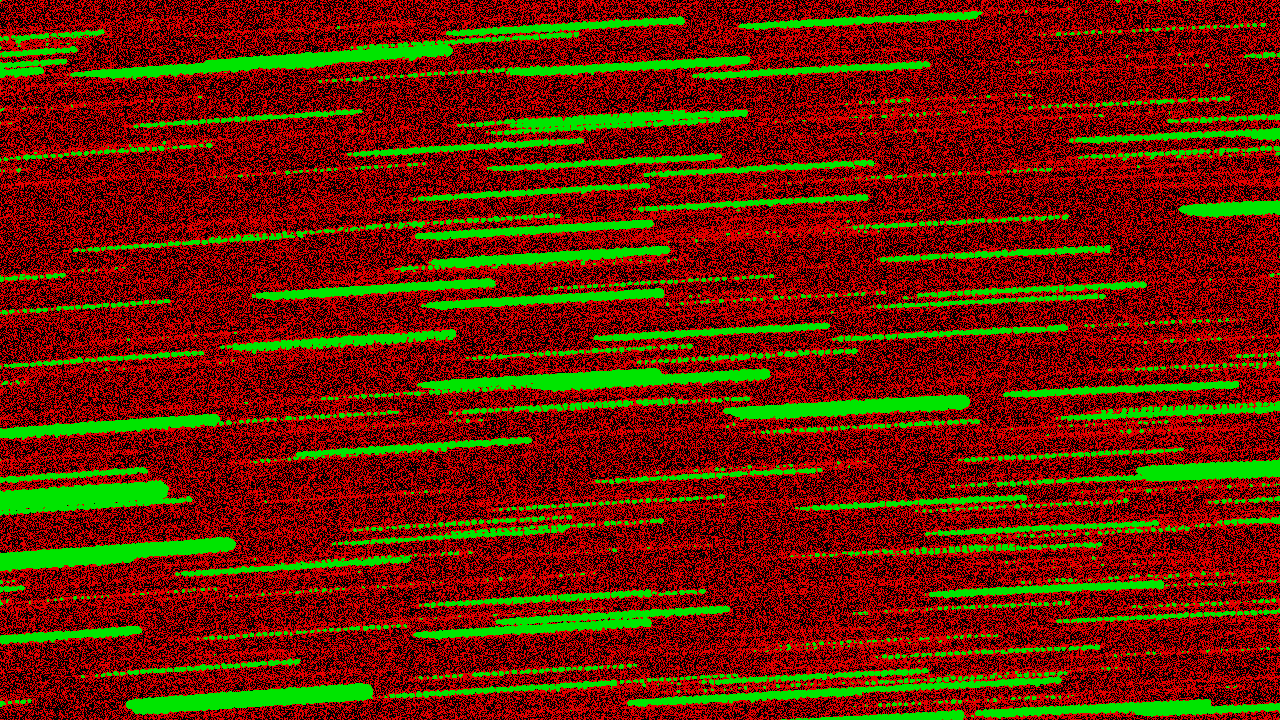}
    
    \caption{Sidereal motion in the \texttt{static} sequence. \textbf{Left}: The event stream visualized for the 3-minute long exposure \textbf{Middle)}: Median filtering to isolate tracks from stars, \textbf{Right)}: Combined visualization for noise and star events.}
    \label{fig:sidereal}
\end{figure*}

\begin{figure*}[htbp]
\centering
\begin{subfigure}{\textwidth}
\includegraphics[width=0.33\textwidth]{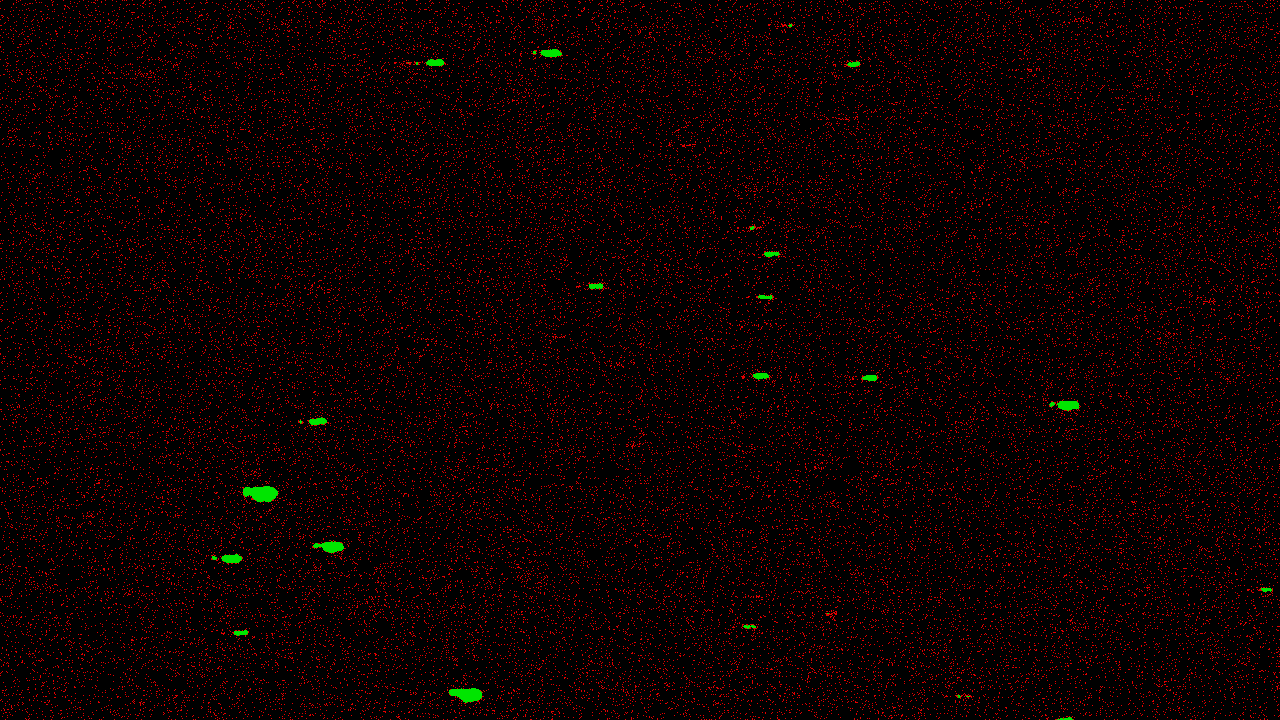} \includegraphics[width=0.33\textwidth]{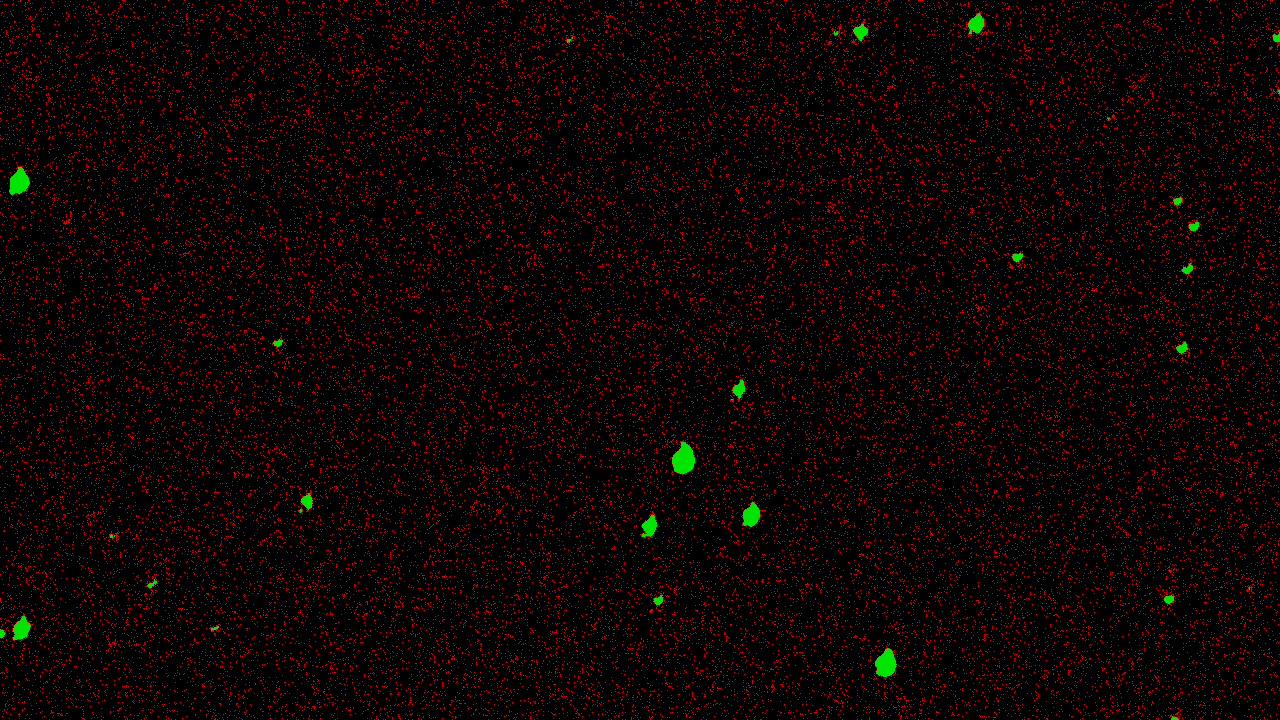} \includegraphics[width=0.33\textwidth]{{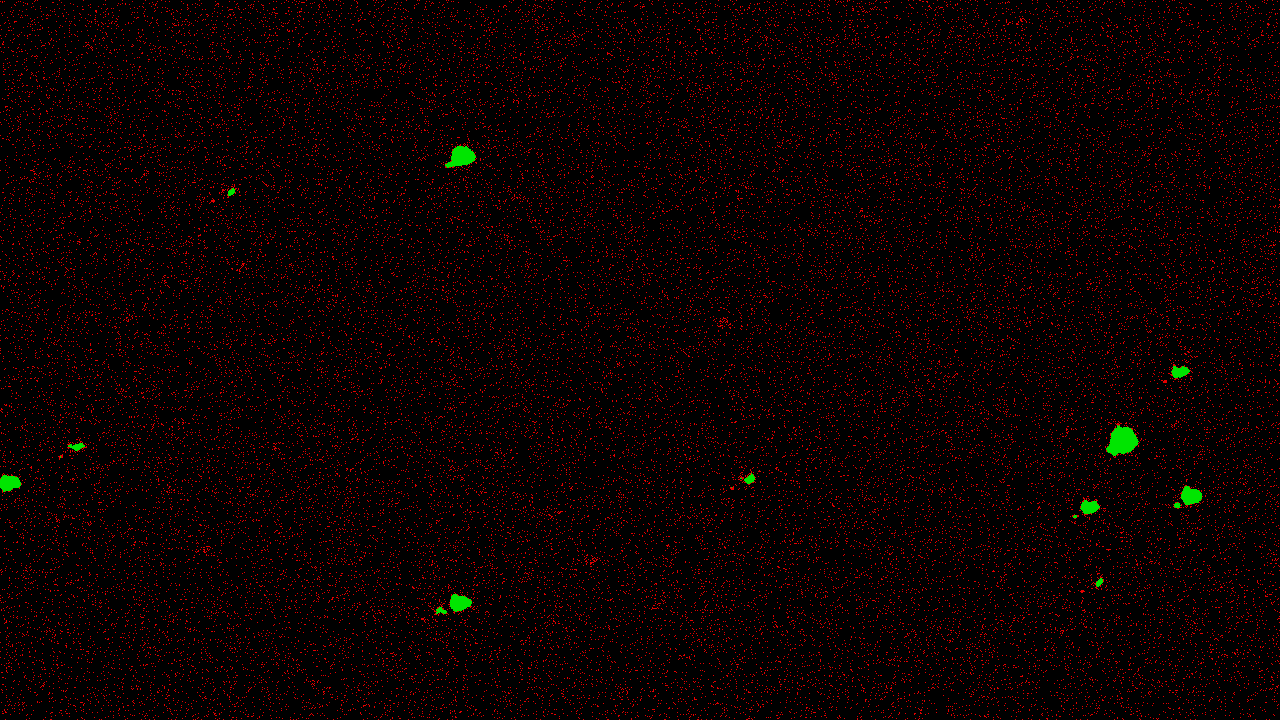}}
\caption{\texttt{slow} sequences corresponding to the 0-30Hz frequency range }
\end{subfigure}\\
\begin{subfigure}{\textwidth}
\includegraphics[width=0.33\textwidth]{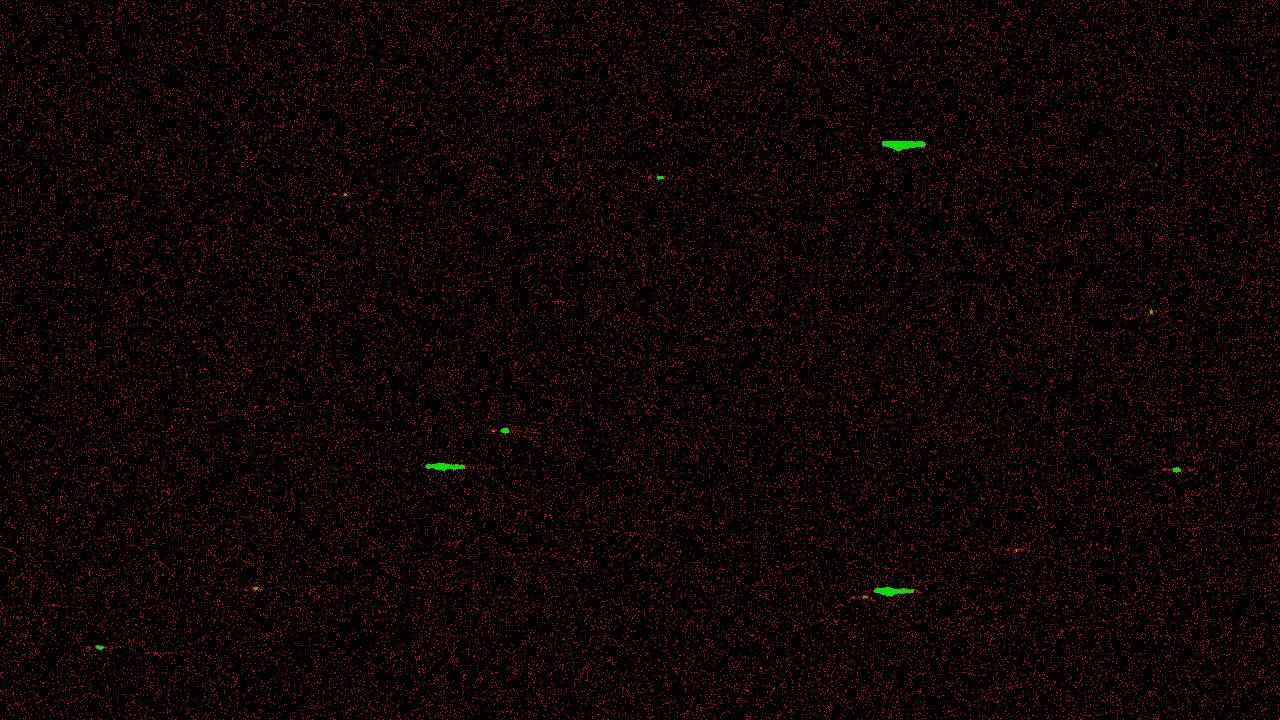} \includegraphics[width=0.33\textwidth]{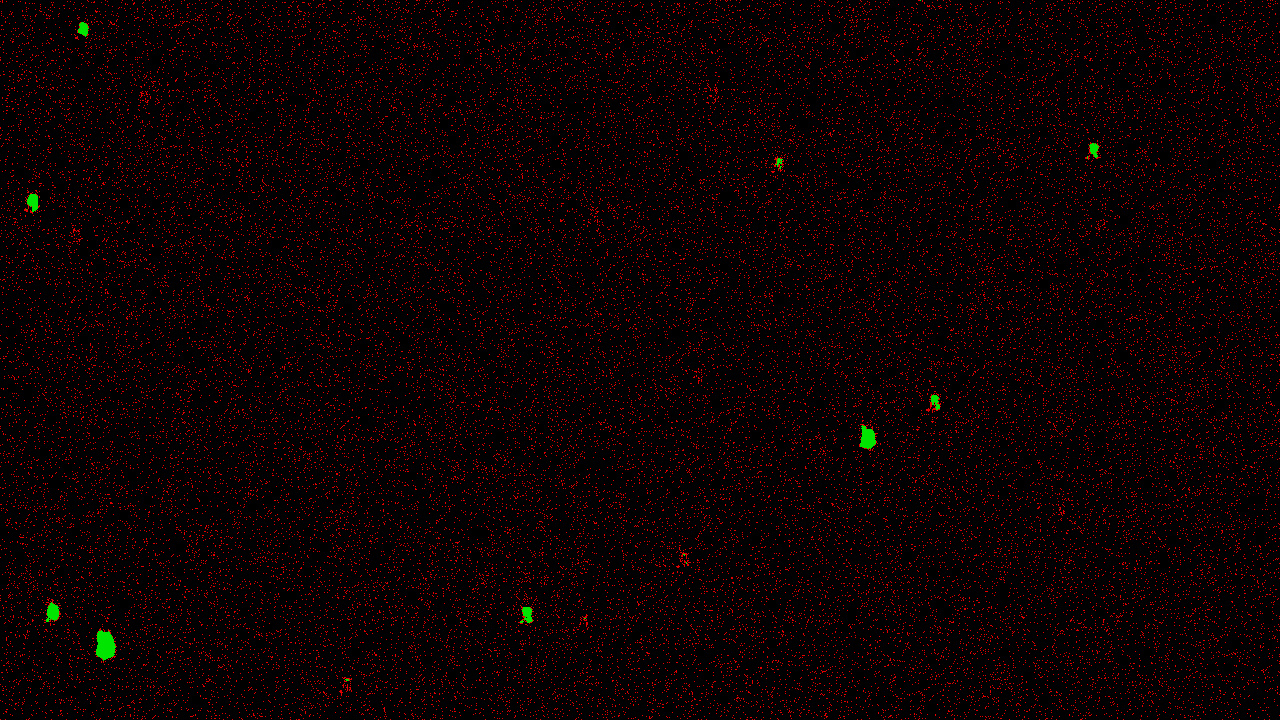} \includegraphics[width=0.33\textwidth]{{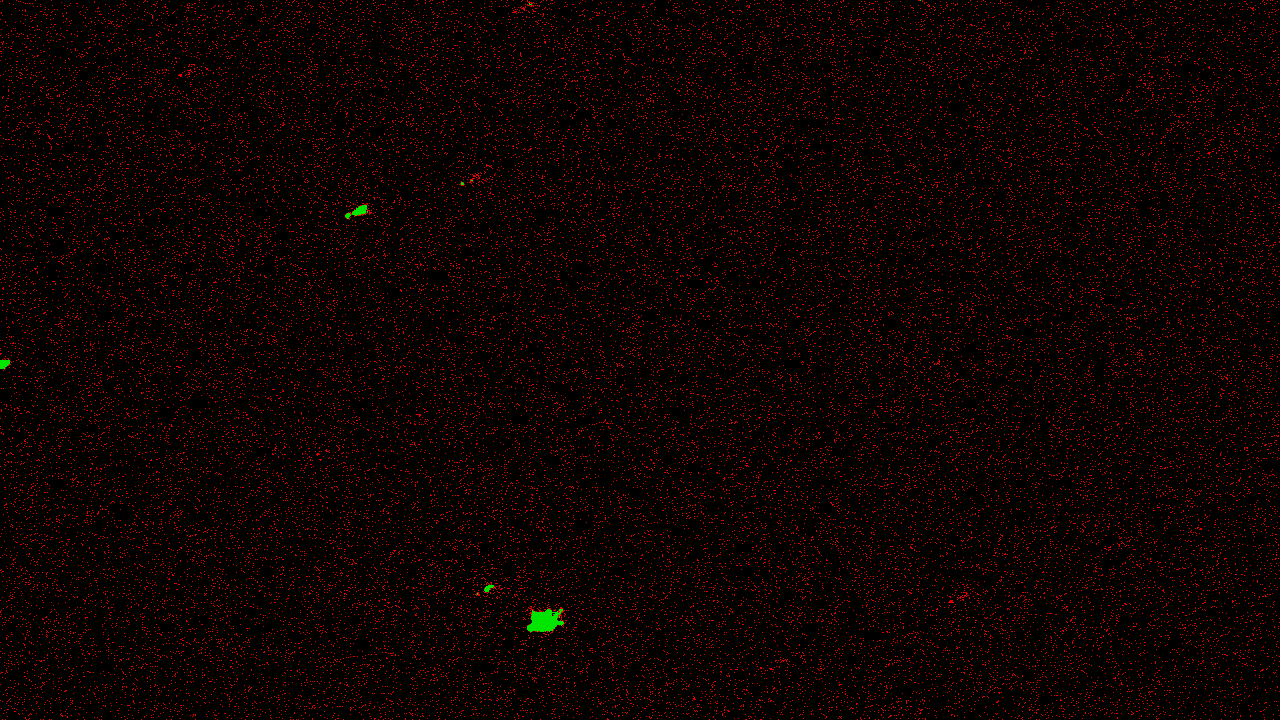}}
\caption{\texttt{medium} sequences corresponding to the 30-100Hz frequency range}
\end{subfigure}\\
\begin{subfigure}{\textwidth}
\includegraphics[width=0.33\textwidth]{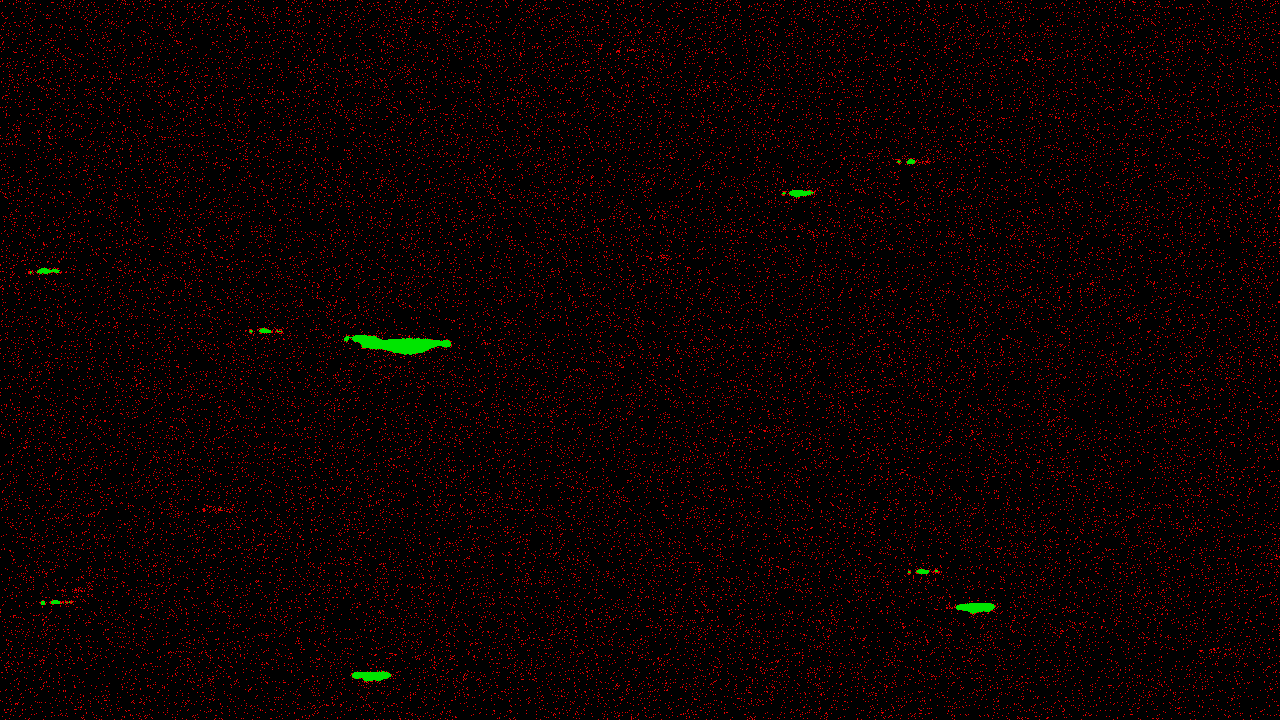}  \includegraphics[width=0.33\textwidth]{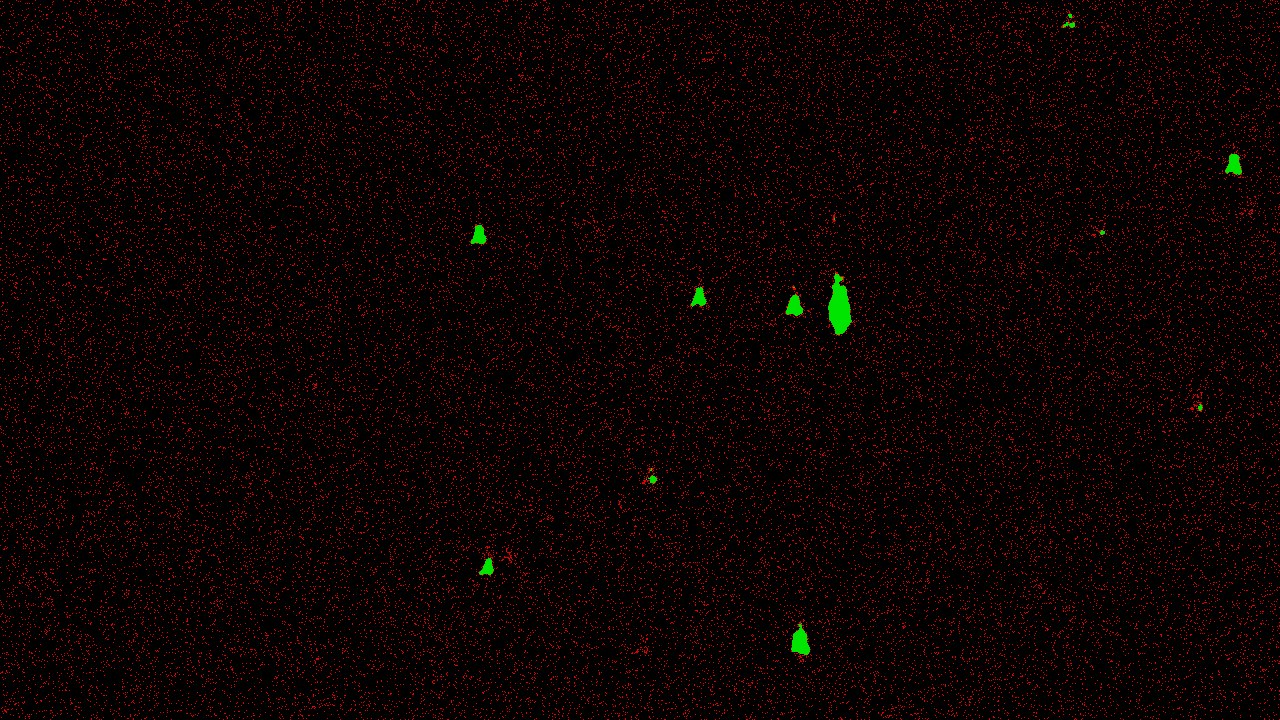} \includegraphics[width=0.33\textwidth]{{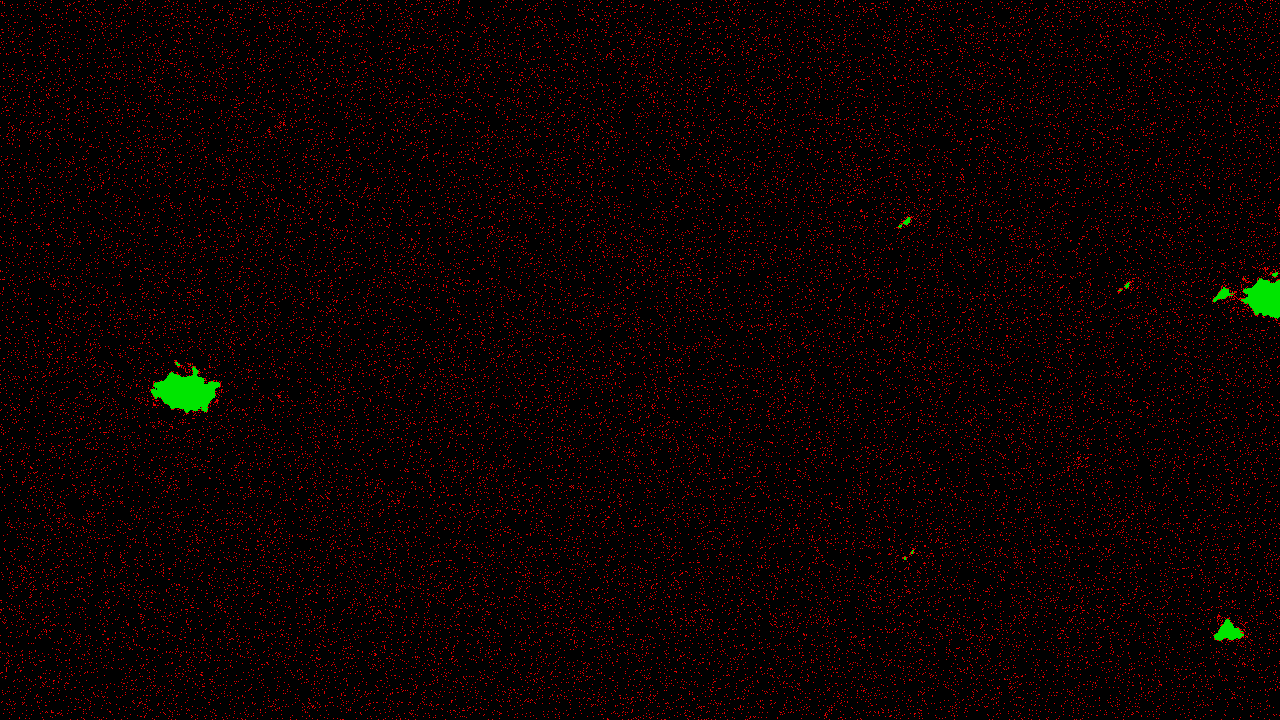}}
\caption{\texttt{fast} sequences corresponding to the 100-200Hz frequency range}
\end{subfigure}\\
% \end{tabular}
\caption{Visualization of the first second of data at the start of \texttt{sequence 17}. Columns represent motion along the first axis, second axis and both axes of the piezoelectric actuator respectively. Events belonging to a star are marked in green and noise is represented in red. (Better seen digitally)}
\label{fig:speed_frequency}
\end{figure*}

\begin{figure*}[t]
    \centering
    \includegraphics[width=0.24\linewidth]{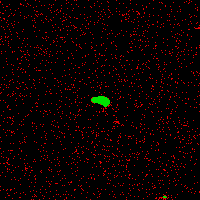}
    \includegraphics[width=0.24\linewidth]{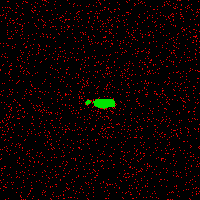}
    \includegraphics[width=0.24\linewidth]{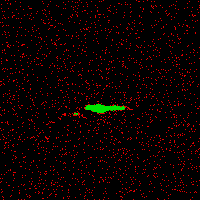}
    \includegraphics[width=0.24\linewidth]{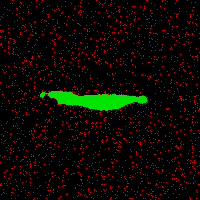}
    \caption{Comparison of induced motion on the sensor plane due to jitter at various frequencies \textbf{(L-R)} no jitter (sidereal motion and atmospheric perturbations), slow (0-30Hz), medium (30-100Hz), and fast (100-200Hz). Events from stars (green) and noise (red). Each crop corresponds to a 200x200 pixel region in the sensor.}
    \label{fig:motion-full-image}
\end{figure*}

\begin{figure*}
\centering
\includegraphics[width=0.24\textwidth]{fast_AXIS1_rgb_crop.png}    \includegraphics[width=0.24\textwidth]{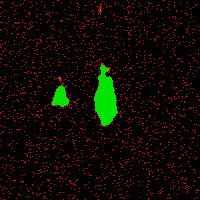}
\includegraphics[width=0.24\textwidth]{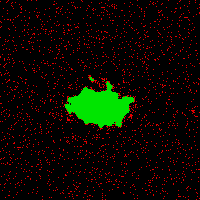}
\caption{Comparison of jitter along various axis of \texttt{Sequence 17-fast}. Events from stars (green) and noise (red). Each crop corresponds to a 200x200 pixel region in the sensor.}
\label{fig:motion-axes}
\end{figure*}

\begin{figure*}
    \centering
    \includegraphics[width=0.32\linewidth]{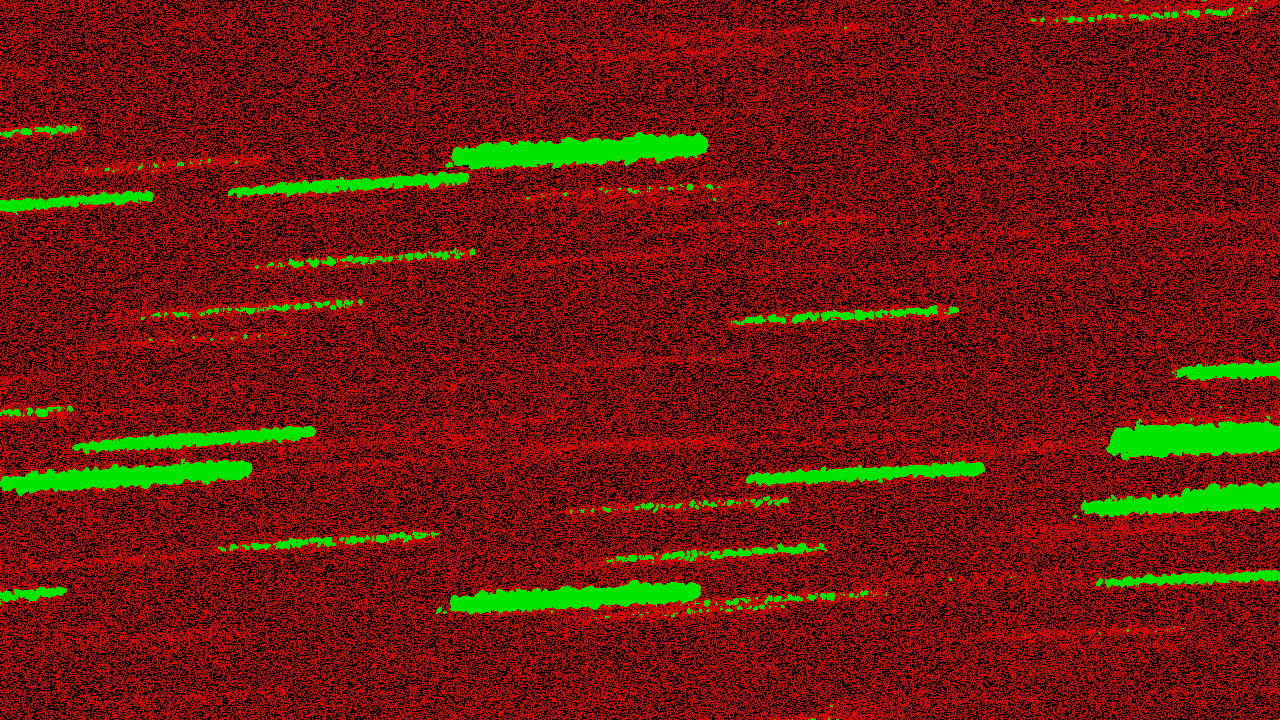}
    \includegraphics[width=0.32\linewidth]{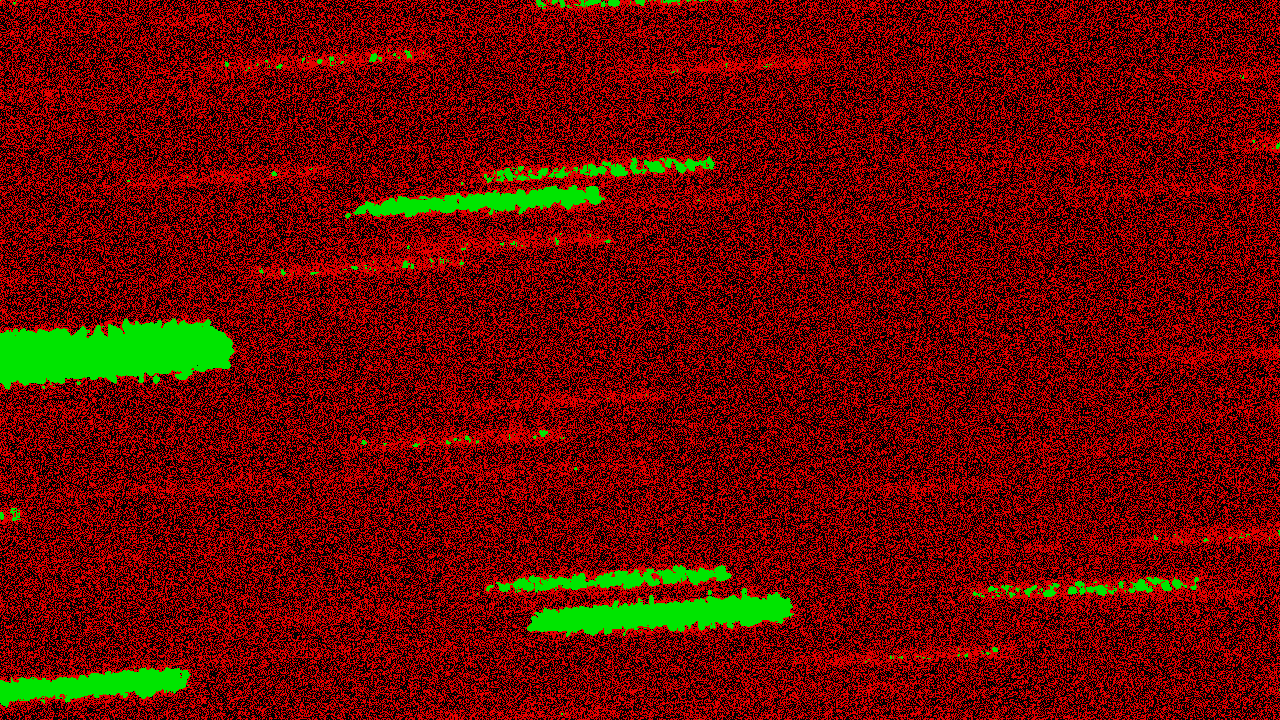}
    \includegraphics[width=0.32\linewidth]{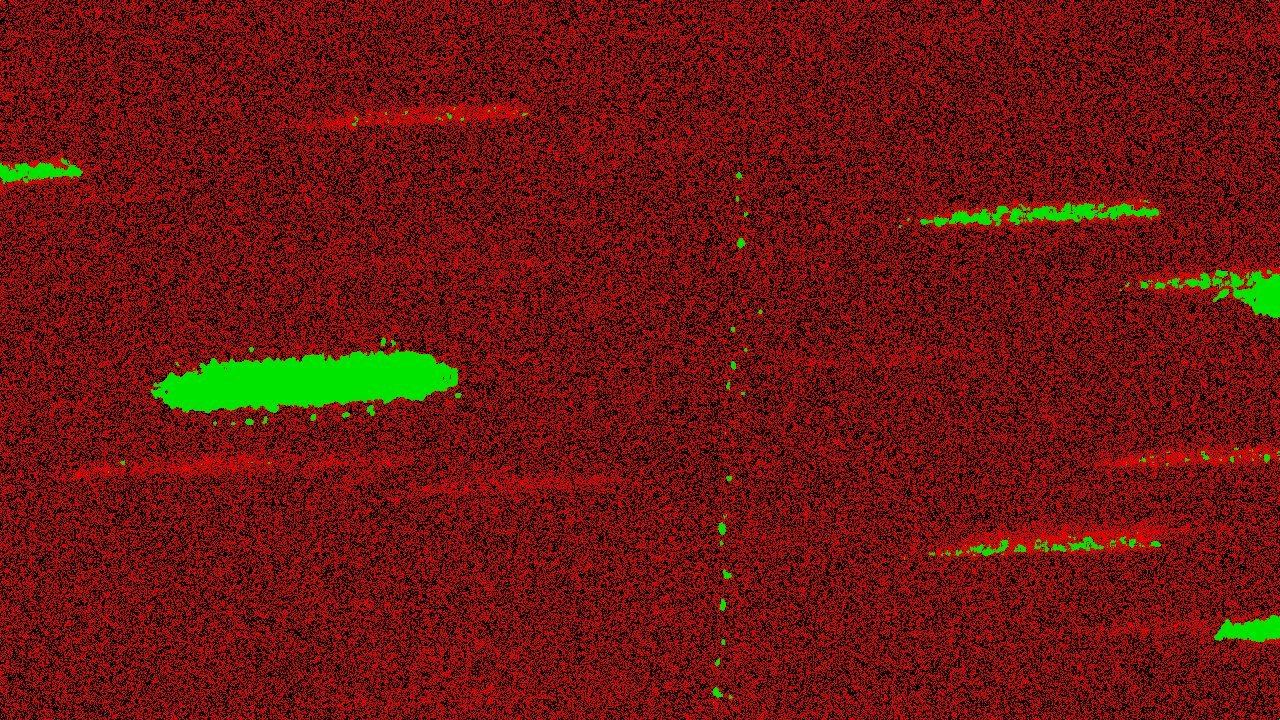}
    \caption{Effect of two axes jitter over the sequence. Star tracks are spread out due to vertical jitter (AXIS 2) as most of the horizontal jitter (AXIS 1) is along the direction of motion. \textbf{L-R)}: \texttt{slow}, \texttt{medium}, and \texttt{fast} sequences.}
    \label{fig:motion-full-sequence}
\end{figure*}

Figure \ref{fig:sidereal} visualizes the \texttt{static} reference sequence, highlighting the subtle sidereal motion captured over a 3-minute exposure. 
Figure \ref{fig:speed_frequency} visualizes \texttt{sequence 17} from the dataset where events associated with stars are highlighted in green, while noise events are depicted in red and Figure \ref{fig:motion-full-image} illustrates the effect of different jitter frequencies on the observed star field, contrasting \texttt{static} conditions with \texttt{slow}, \texttt{medium}, and \texttt{fast} jitter along a single axis.  The impact of axis-specific jitter is shown in Figure \ref{fig:motion-axes}, where \texttt{fast} jitter is applied along the first axis, second axis, and both axes, demonstrating the resulting event patterns.
The cumulative effect of two-axis jitter over an entire sequence is depicted in Figure \ref{fig:motion-full-sequence}, illustrating how star tracks are dispersed due to vertical jitter, while horizontal jitter, aligned with the direction of motion, results in more elongated tracks. 

\begin{algorithm}[t]
\caption{MACRO\_CIRCULAR\_QUEUE ($N$,$amplitude$, $delay$)}
\begin{algorithmic}[1]
\STATE $StepCount \gets 0$
\FOR{$i$ = 1 \dots $N$}
\STATE Move {$axis1$ by $amplitude$}
\STATE Move {$axis2$ by $amplitude$}
\STATE Delay {$delay$}
\STATE $output1 \gets LoopCount$
\STATE $output2 \gets AXIS1\_POS$
\STATE $StepCount \gets StepCount+1$
\STATE $output3 \gets StepCount$
\STATE $output4 \gets AXIS2\_POS$
\STATE $StepCount \gets StepCount+1$
\STATE $output5 \gets StepCount$
\STATE Move {$axis1$ by -$amplitude$}
\STATE Move {$axis2$ by -$amplitude$}
\STATE Delay {$delay$}
\STATE $output6 \gets AXIS1\_POS$
\STATE $StepCount \gets StepCount+1$
\STATE $output7 \gets StepCount$
\STATE $output8 \gets AXIS2\_POS$
\STATE $StepCount \gets StepCount+1$
\STATE $output9 \gets StepCount$
\STATE $LoopCount \gets LoopCount+1$
\ENDFOR
\end{algorithmic}
\label{alg:circular_queue}
\end{algorithm}

\begin{figure*}
    \centering
    \includegraphics[width=0.8\linewidth]{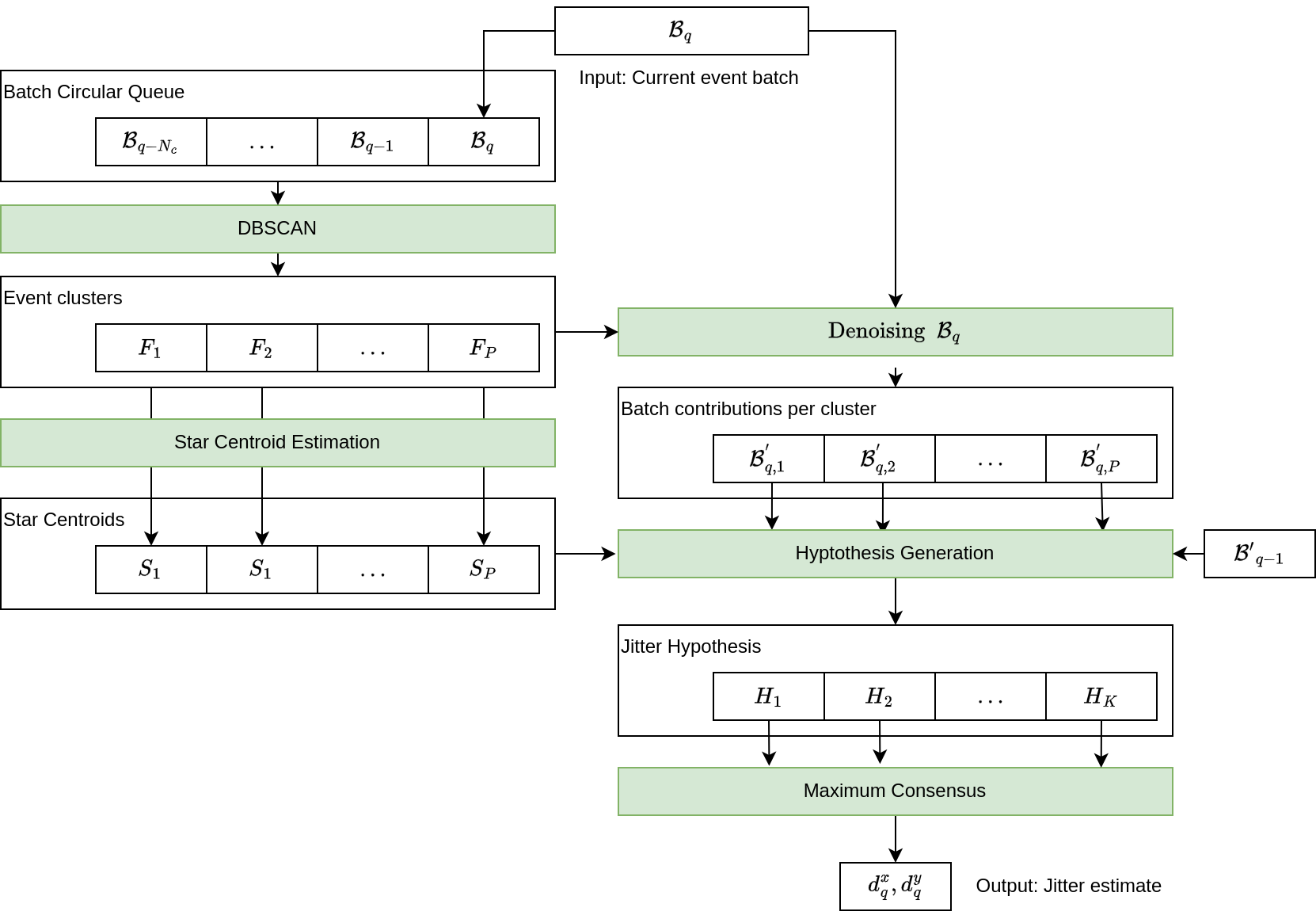}
    \caption{Overview of the vibration recovery algorithm. Coloured boxes represent operations in the pipeline. Each batch of data $\mathcal{B}_{q}$ generates an estimated of the vibration $(d_q^x, d_q^y)$ that occurred during its time span, see text for details.}
    \label{fig:flow}
\end{figure*}
\begin{figure}
    \centering
    \includegraphics[width=\linewidth]{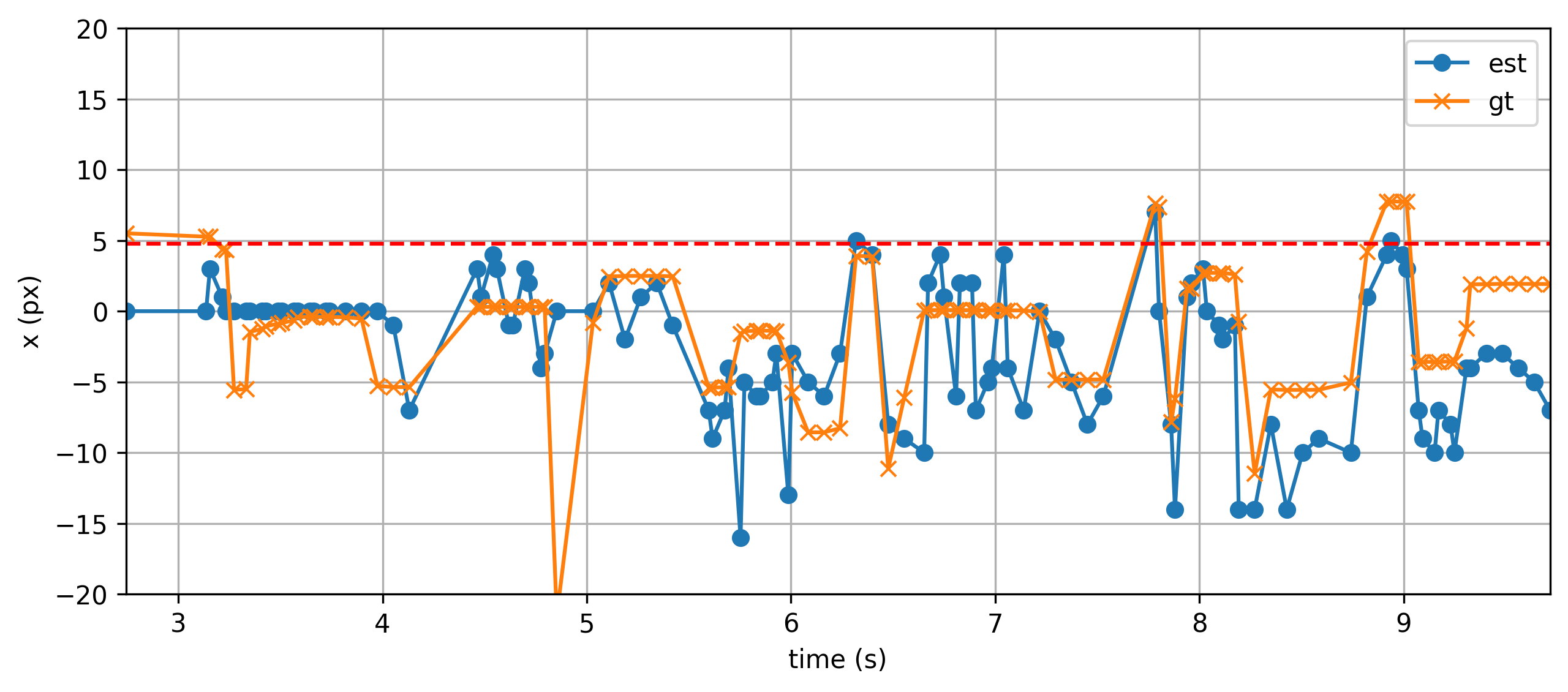}\\
    \includegraphics[width=\linewidth]{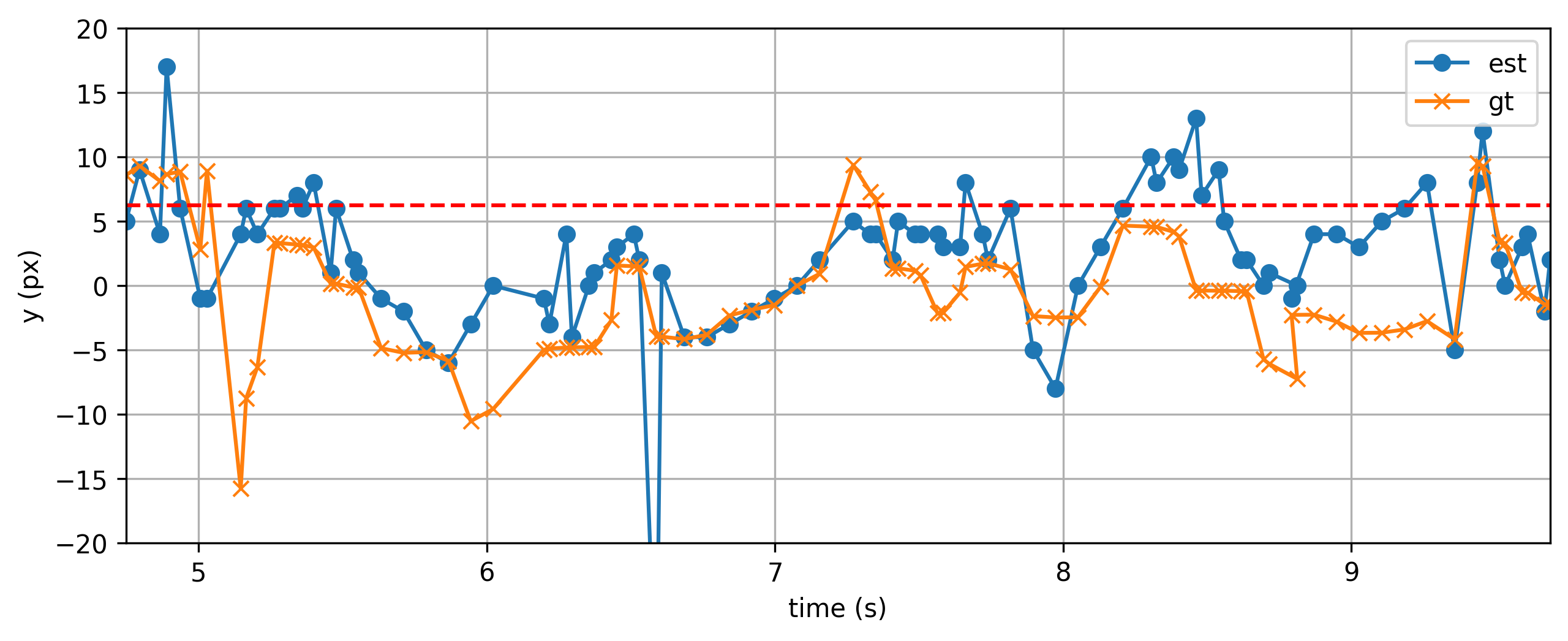 }\\
    
    \caption{Ground Truth and estimated jitter for \texttt{Axis1} and \texttt{Axis2} for \texttt{Sequence 17 - BothAxis - Slow} configuration. Red dotted line marks the RMSE error between the two. (First 10 seconds shown for visibility).}
    \label{fig:x_y_plot}
\end{figure}

% \begin{figure*}
%     \centering
%     \includegraphics[width=0.45\linewidth]{xyt.png}
%     \includegraphics[width=0.45\linewidth]{xt.png}\\
%     \includegraphics[width=0.45\linewidth]{yt.png}
%     \caption{\hl{MAKE THIS TITLE FIGURE: Jitter in the XYT space.}}
%     \label{fig:enter-label}
% \end{figure*}

\section{Vibration Recovery Algorithm}
\label{sec:algorithm}

Event-based cameras generate asynchronous streams of events, $\mathcal{E} = \{e_k\}_{k=1}^{N}$, where each event $e_k = (x_k, y_k, t_k, p_k)$. Here, $(x_k, y_k)$ are the pixel coordinates of the event, $t_k$ is its timestamp, and $p_k \in \{-1, +1\}$ indicates the polarity of the brightness change: $p_k = +1$ signifies an increase in brightness, and $p_k = -1$ indicates a decrease.  Since a single event provides insufficient information for jitter recovery, we aggregate events over short time intervals to form event ``batches''.

\subsection{Event Batching}
We partition the incoming event stream into non-overlapping batches. Each batch, $\mathcal{B}_q$, spans a duration of $t_{\text{batch}}$ seconds.  The $q$-th batch is defined as:
\begin{equation}
\mathcal{B}_q = \{e_k \mid t_k \in [q \cdot t_{\text{batch}}, (q + 1) \cdot t_{\text{batch}})\}
\end{equation}
\noindent
assuming $t_0 = 0$ (the start of the event stream).  Thus, $\mathcal{B}_0$ contains events from the first $t_{\text{batch}}$ seconds. The value of $t_{\text{batch}}$ is determined by the maximum jitter frequency (Sec.~\ref{sec:frequency_band_selection}).  This batching process preserves the temporal information for each event, as we do not convert events into frames.

\subsection{Clustering and Noise Removal}

The star field observed by a star tracker is inherently sparse, with few pixels illuminated by star photons.  This sparsity, combined with spurious noise events (from electronic interference, thermal fluctuations, or ambient light), lowers the signal-to-noise ratio (SNR). Effective jitter recovery requires separating star-generated events (signal) from noise.

We employ a clustering-based approach, leveraging the assumption that noise is random and unstructured, while star-generated events under jitter are spatiotemporally clustered.  Directly applying clustering to a single batch, $\mathcal{B}_q$, is ineffective due to its short duration ($t_{\text{batch}}$). Therefore, we maintain a circular queue of $N_c$ past batches.  For the current batch, $\mathcal{B}_q$, we construct a combined event set, $\mathcal{C}_q$:
\begin{equation}
\mathcal{C}_q = \bigcup_{p = q - N_c}^{q} \mathcal{B}_p
\end{equation}

We apply Density-Based Spatial Clustering of Applications with Noise (DBSCAN)~\cite{ester1996density} to $\mathcal{C}_q$ to identify event clusters corresponding to individual stars. DBSCAN finds arbitrary shaped clusters and is robust to noise. It outputs a set of $P$ clusters,  $\{F_1, F_2, ..., F_P\}$, where each $F_p$ ideally represents events originating from a single star.

\subsection{Star Centroid Estimation}

Given the clusters, $\{F_1, F_2, ..., F_P\}$, we estimate the centroid of each star using the last $N_c$ batches. The star's center is not directly visible in the event stream; only moving edges generate events.  However, the star oscillates around a mean position, which we recover as its centroid.

For each cluster $F_p$, we consider the events $\{(x_j^p, y_j^p, t_j^p)\} \in F_p$.  We estimate two lines, representing the x-coordinate vs. time (XT) and y-coordinate vs. time (YT) relationships:
\begin{equation}
x_j^p = m_x^p t_j^p + c_x^p
\end{equation}
and
\begin{equation}
y_j^p = m_y^p t_j^p + c_y^p
\end{equation}
\noindent
using the method of normals~\cite{golub2013matrix} to estimate the parameters $(m_x^p, c_x^p)$ and $(m_y^p, c_y^p)$, which describe the linear trend of the star's motion in x and y over time. We evaluate these lines at the current time, $t_q = (q+1) \cdot t_{\text{batch}}$, to obtain the centroid, $S_p$, of the $p$-th star:
\begin{equation}
S_p = (m_x^p t_q + c_x^p, m_y^p t_q + c_y^p)
\end{equation}

Algorithm~\ref{alg:centroid_estimation} summarizes this process.

\begin{algorithm}
\caption{Star Centroid Estimation per Cluster}
\label{alg:centroid_estimation}
\begin{algorithmic}[1]
\REQUIRE Clusters $\{F_p\}_{p=1}^P$, Batch duration $t_{\text{batch}}$
\ENSURE Estimated star centroid $S_p$ for each $p$

\FOR{each cluster $F_p$}
    \STATE Fit XT line: $x_j^p = m_x^p t_j^p + c_x^p$ using the method of normals
    \STATE Fit YT line: $y_j^p = m_y^p t_j^p + c_y^p$ using the method of normals
    \STATE $t_q \gets (q+1) \cdot t_{\text{batch}}$
    \STATE $S_p \gets (m_x^p t_q + c_x^p, m_y^p t_q + c_y^p)$
\ENDFOR
\end{algorithmic}
\end{algorithm}

\subsection{Jitter Hypothesis Generation}

The clusters contain events from the last $N_c$ batches. We extract events belonging to the current batch, $\mathcal{B}'_{q,p}$ as the intersection of $\mathcal{B}_q$ with each cluster $F_p$:
\begin{equation}
\mathcal{B}'_{q,p} = \mathcal{B}_q \cap F_p, \quad p = 1 \dots P
\end{equation}
\noindent
where $\mathcal{B}'_{q,p}$ ideally contains only events from the $p$-th star in the current batch.  We now have the star's centroid, $S_p$, and its associated events in the current batch, $\mathcal{B}'_{q,p}$.

To recover the jitter, we compare events in $\mathcal{B}'_{q,p}$ with those in the previous batch's corresponding set, $\mathcal{B}'_{q-1,p}$.  To mitigate outliers, we define a support set, $W_{q,p}$, containing events in $\mathcal{B}'_{q,p}$ within a radius, $r$, of the centroid, $S_p$:

\begin{equation}
W_{q,p} = \{e_k \in \mathcal{B}'_{q,p} \mid \|(x_k,y_k) - S_p\| < r\}
\end{equation}

\noindent
where $r$ represents the maximum expected jitter amplitude, determined by the piezoelectric stage's maximum displacement and the motion-to-pixel calibration.

We compare $W_{q,p}$ against $W_{q-1,p}$ under a set of motion hypotheses, $\mathbf{H}_x$ and $\mathbf{H}_y$, representing possible x and y displacements.  We select the hypothesis with maximum support:
\begin{equation}
h_x^{*}, h_y^{*} = \arg\max_{(h_x, h_y)} \sum_{e_n \in W_{q-1,p}} I[(x_n  + h_x, y_n + h_y) \in W_{q,p}]
\end{equation}
\noindent
where $I[\cdot]$ is an indicator function: $I[\text{condition}] = 1$ if the condition is true, and 0 otherwise.  This finds the displacement $(h_x, h_y)$ that maximizes the number of events in the current support set ($W_{q,p}$) with corresponding events in the previous support set ($W_{q-1,p}$) after applying the displacement. We ignore event timestamps and polarities in this matching, focusing on spatial proximity. The resulting $(h_x^*, h_y^*)$ represents the estimated jitter between batches $q-1$ and $q$.
%

% For two consecutive event batches, $\mathcal{E}^{p}_t$ and $\mathcal{E}^{p}_{t+1}$, we formulate the motion recovery as a maximum consensus problem. Each event in $\mathcal{E}^{p}_t$ is displaced by a possible motion hypothesis $\delta x, \delta y$ and compared position of events in $\mathcal{E}^{p}_{t+1}$. The motion of hypothesis that leads to maximum number of common events is selected as the incremental motion between the two event frames. 

% \subsection{Keyframes based tracking}
% The frame-to-frame method for motion recovery is prone to drift: the overall motion estimate is directly effected by a single incorrect motion estimate. For robustness, we adopt a keyframe based approach, wherein, motion is estimated against a fixed frame -- the keyframe -- until a predefined amount of time elapses or amount of motion is observed. 

% \begin{algorithm}
% \caption{Event-Based Jitter estimation}
% \begin{algorithmic}[1]
% \STATE \textbf{Input:} Event batch $\mathcal{E}_t$
% \STATE \textbf{Output:} Jitter estimate at $t$
% % \STATE Initialize $x \leftarrow 0$, $y \leftarrow 0$
% % \STATE Initialize $frameID \leftarrow 0$

% \STATE Apply DBSCAN to get clusters $C_i,~i=1 \dots N$
% %\STATE Detect centroids in segmented event clusters
% \FOR{each $C_i$}
%     \STATE estimate lines in XT and YT
%     \STATE estimate star positions
%     \STATE $S_t \gets$ estimate support from $\mathcal{E}_t$
% \ENDFOR
% \STATE align $S_{t-1}$ and $S_t$

% \end{algorithmic}
% \end{algorithm}

\section{Results}\label{sec:results}
We evaluate the performance of the proposed algorithm on the e-STURT dataset. For the three distinct frequencies ranges in the dataset, we use different $t_{batch}$ parameter~( Sec.~\ref{sec:algorithm}) representing the duration of each event batch. The Nyquist-Shannon sampling theorem~\cite{Oppenheim_Schafer_DSP} dictates that $t_{batch} = 1/(2f_{max})$ where $f_{max}$ is the highest frequency should use for a sequence with the hightest frequency $f_max$. This ensures that the sampling frequency of the algorithm is high enough to recover the highest jitter present in the sequence.

% \begin{table}[!h]
%     \centering
%     \caption{The batch time used in experiments for each frequency range.}
%     \label{tab:settings}
%     \begin{tabular}{c|c|c}
%         Setting &  Frequency Range (Hz) & $t_{batch}$ (ms) \\\hline
%         \texttt{slow} & $0$-$30$ & 16.66 \\\hline
%         \texttt{medium} & $30$-$100$ & 5 \\\hline
%         \texttt{fast} & $100$-$200$ & 2.5 \\        
%     \end{tabular}
    
% \end{table}

\begin{figure*}[htbp] % Placement specifier: here, top, bottom, page
    \centering % Center the whole figure

    % First row as a subfigure
    \begin{subfigure}{\textwidth} % This subfigure spans the text width
        \centering % Center the images within this row's subfigure
        \includegraphics[width=0.32\textwidth]{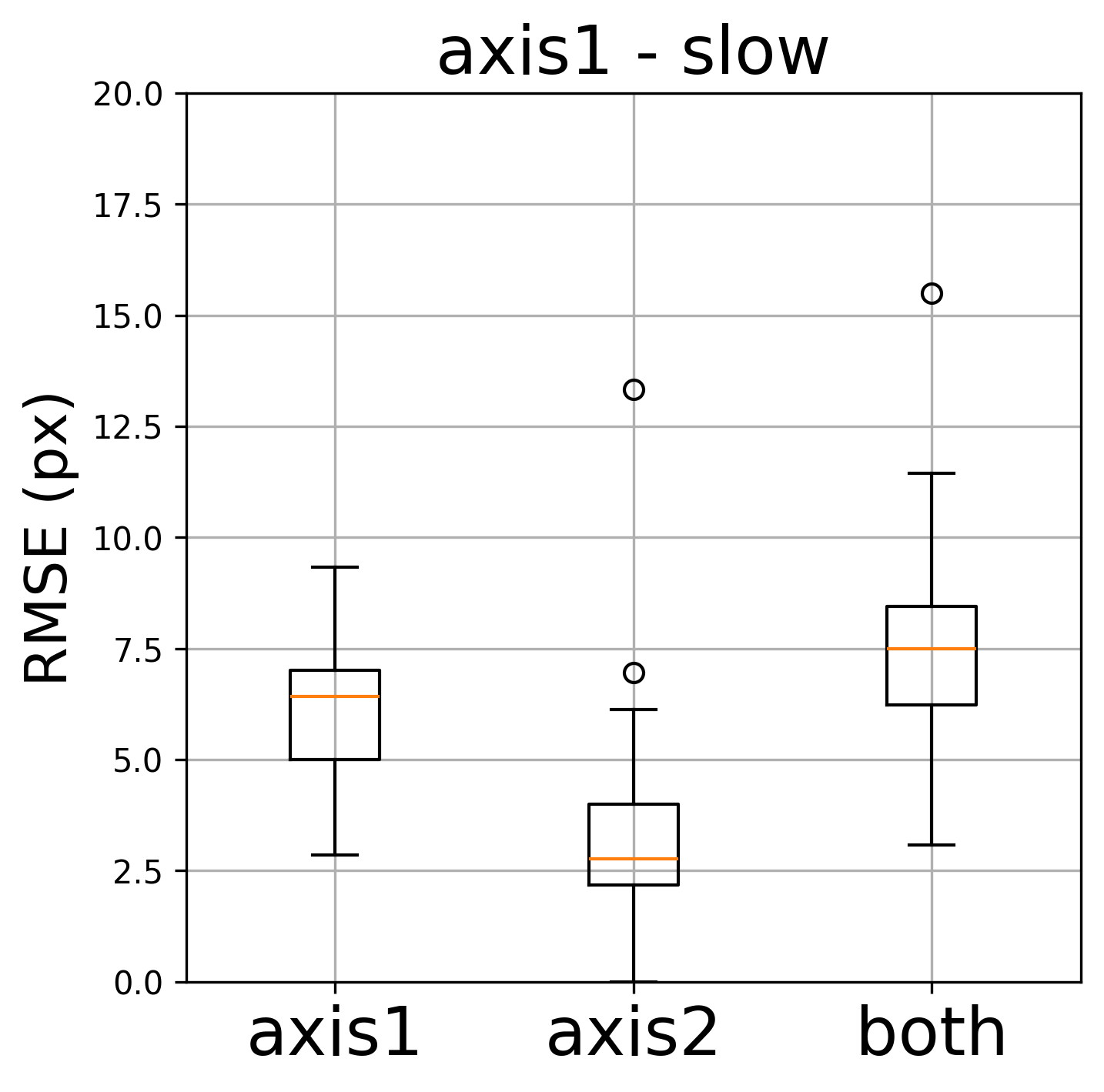} % Image 1, width relative to text width
        \hfill % Add horizontal space between images
        \includegraphics[width=0.32\textwidth]{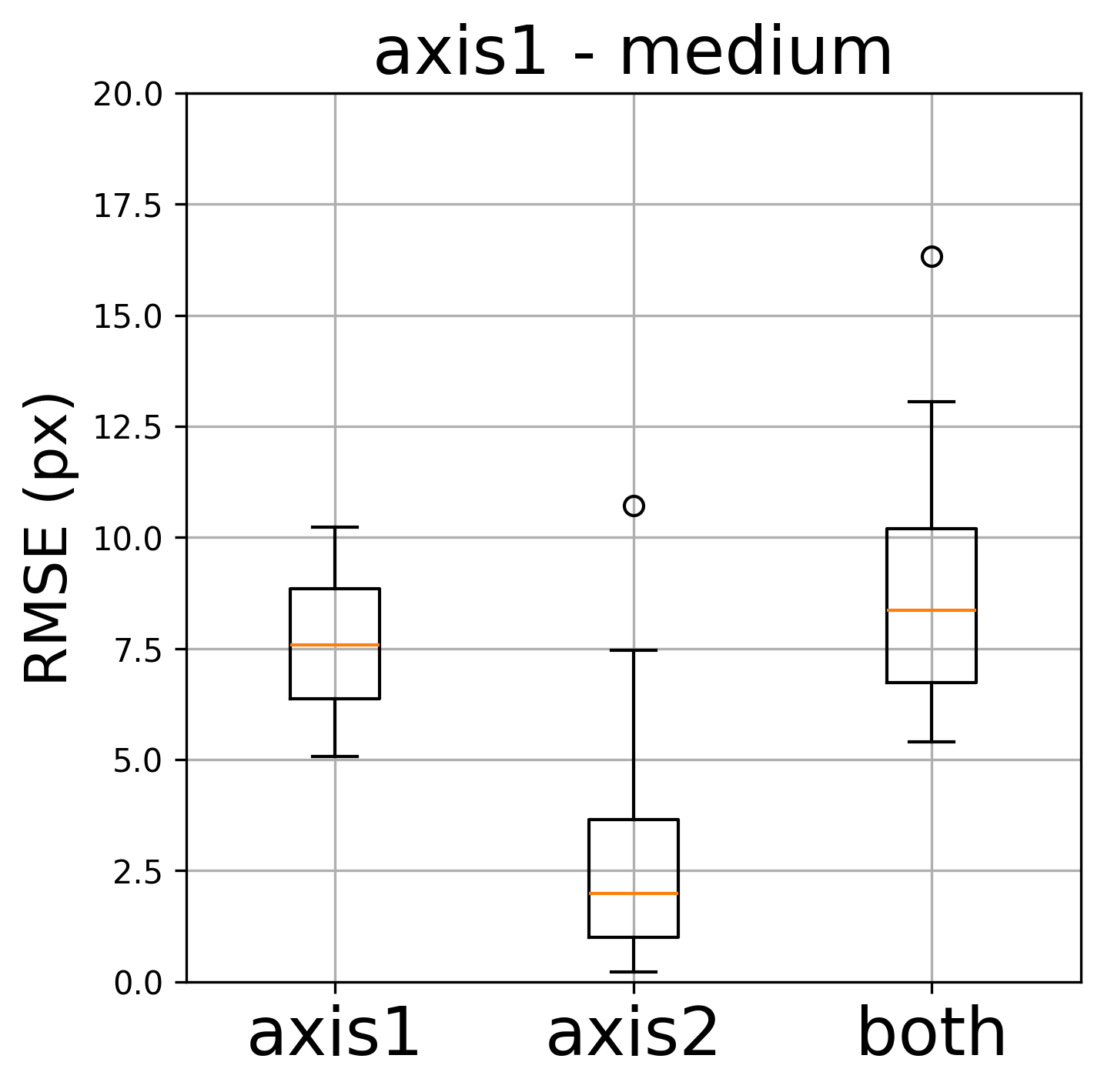} % Image 2, width relative to text width
        \hfill % Add horizontal space between images
        \includegraphics[width=0.32\textwidth]{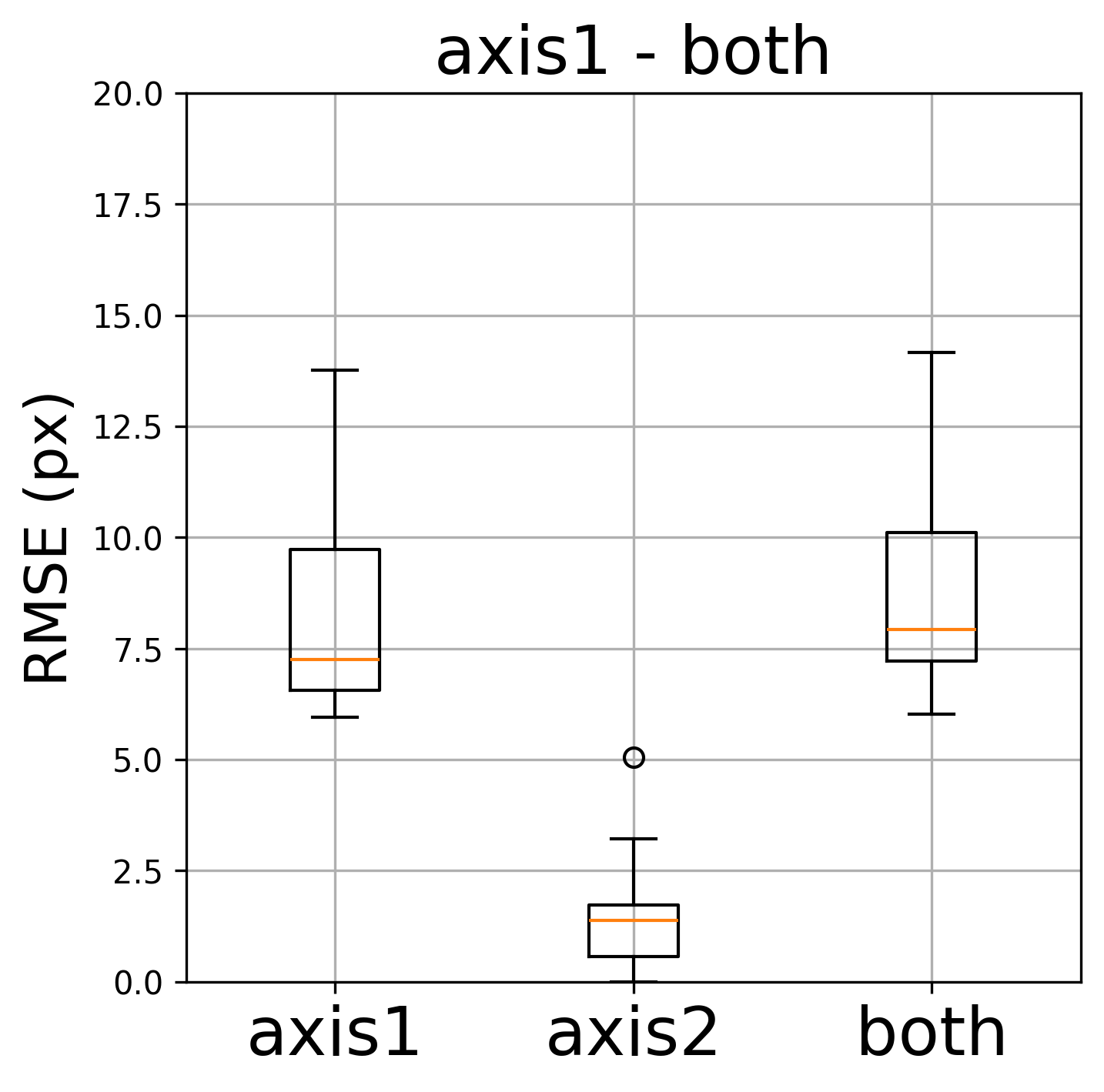} % Image 3, width relative to text width
        \caption{Analysis of jitter recovery for motion along \texttt{axis1}} % Caption for the entire first row
        \label{fig:row1}
    \end{subfigure}

    \bigskip % Add some vertical space between the main row subfigures

    % Second row as a subfigure
    \begin{subfigure}{\textwidth} % This subfigure spans the text width
        \centering % Center the images within this row's subfigure
        \includegraphics[width=0.32\textwidth]{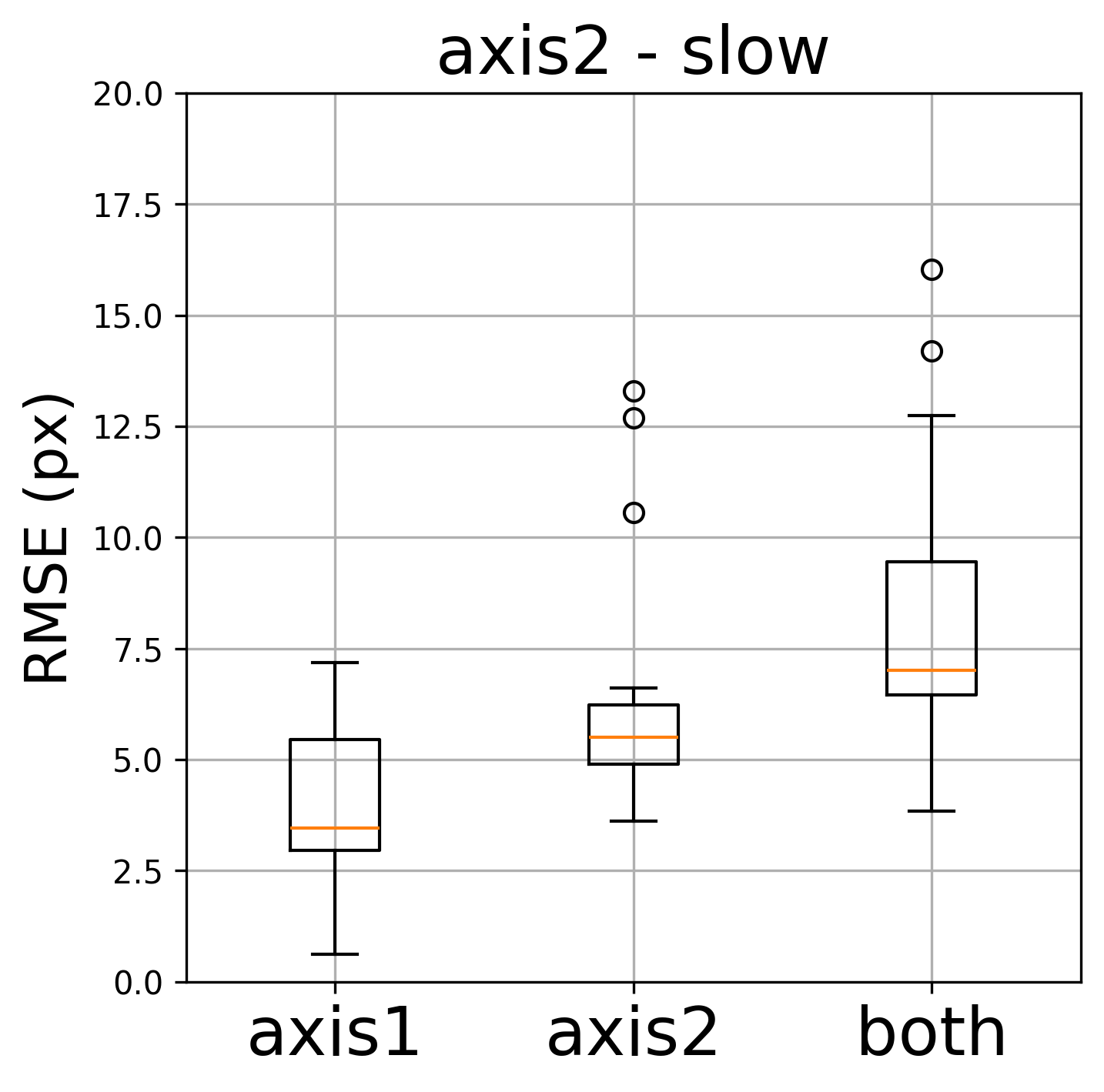} % Image 1, width relative to text width
        \hfill % Add horizontal space between images
        \includegraphics[width=0.32\textwidth]{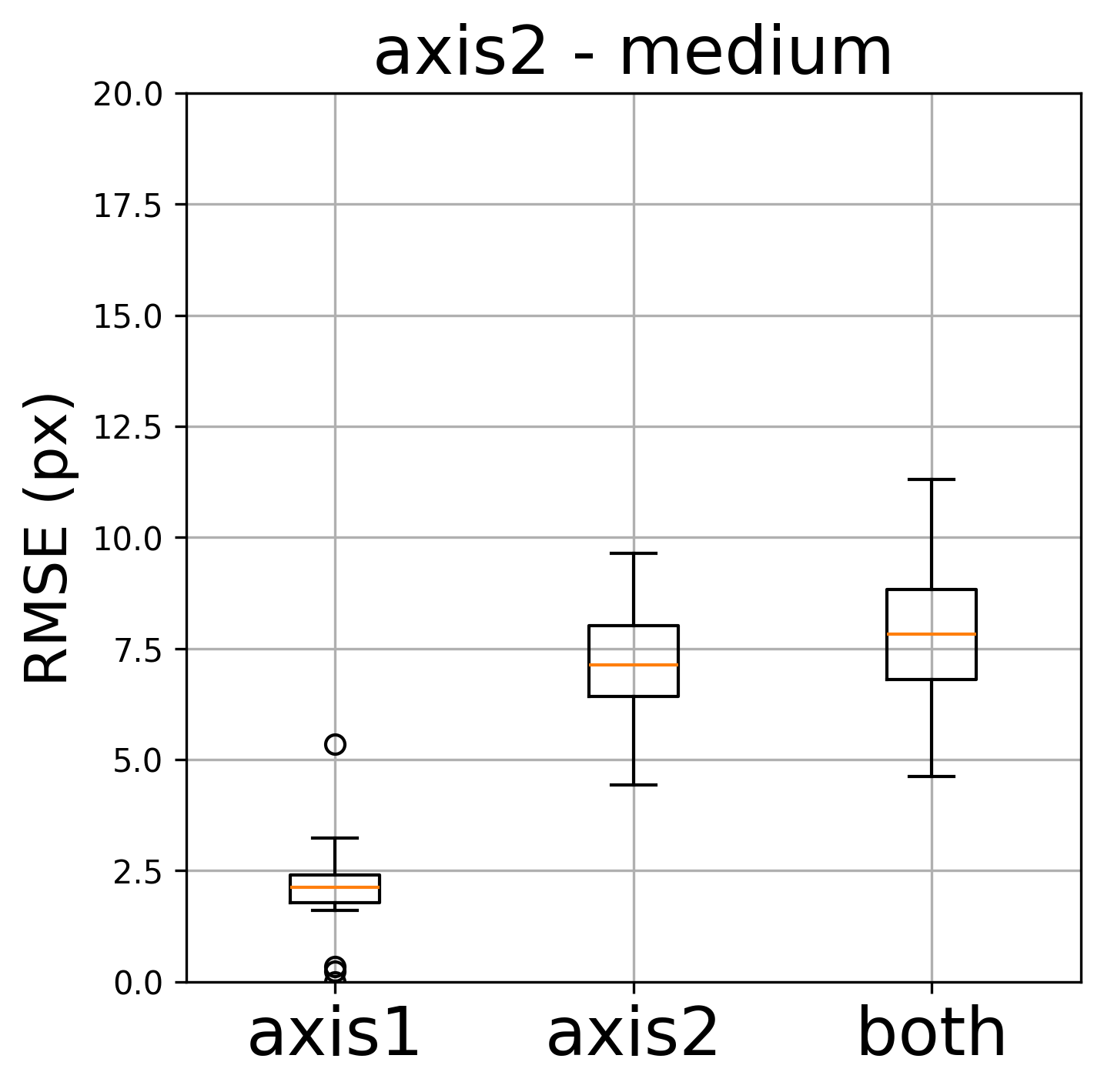} % Image 2, width relative to text width
        \hfill % Add horizontal space between images
        \includegraphics[width=0.32\textwidth]{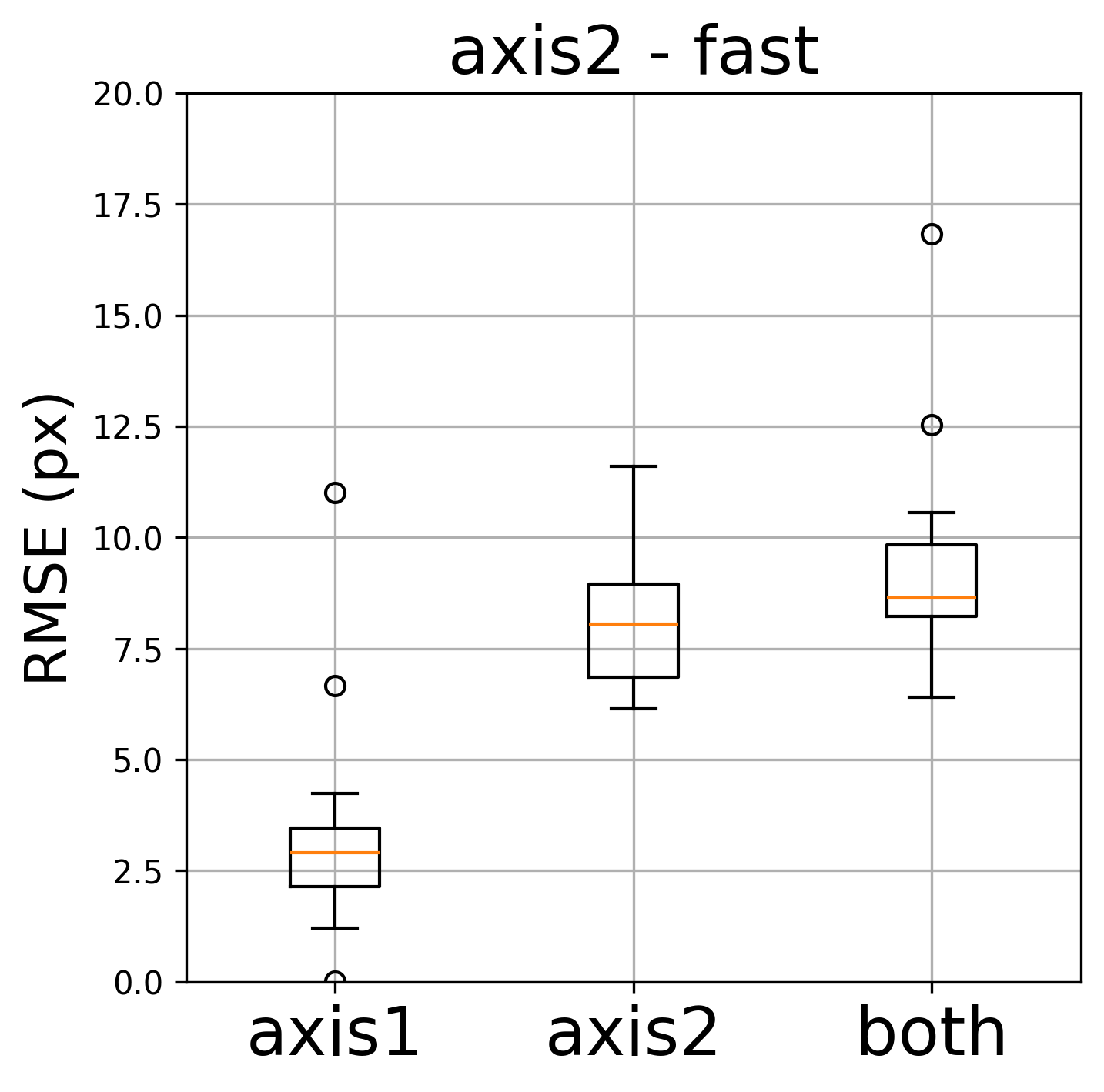} % Image 3, width relative to text width
        \caption{Analysis of jitter recovery for motion along \texttt{axis2}} % Caption for the entire second row
        \label{fig:row2}
    \end{subfigure}

    \bigskip % Add some vertical space between the main row subfigures

    % Third row as a subfigure
    \begin{subfigure}{\textwidth} % This subfigure spans the text width
        \centering % Center the images within this row's subfigure
        \includegraphics[width=0.32\textwidth]{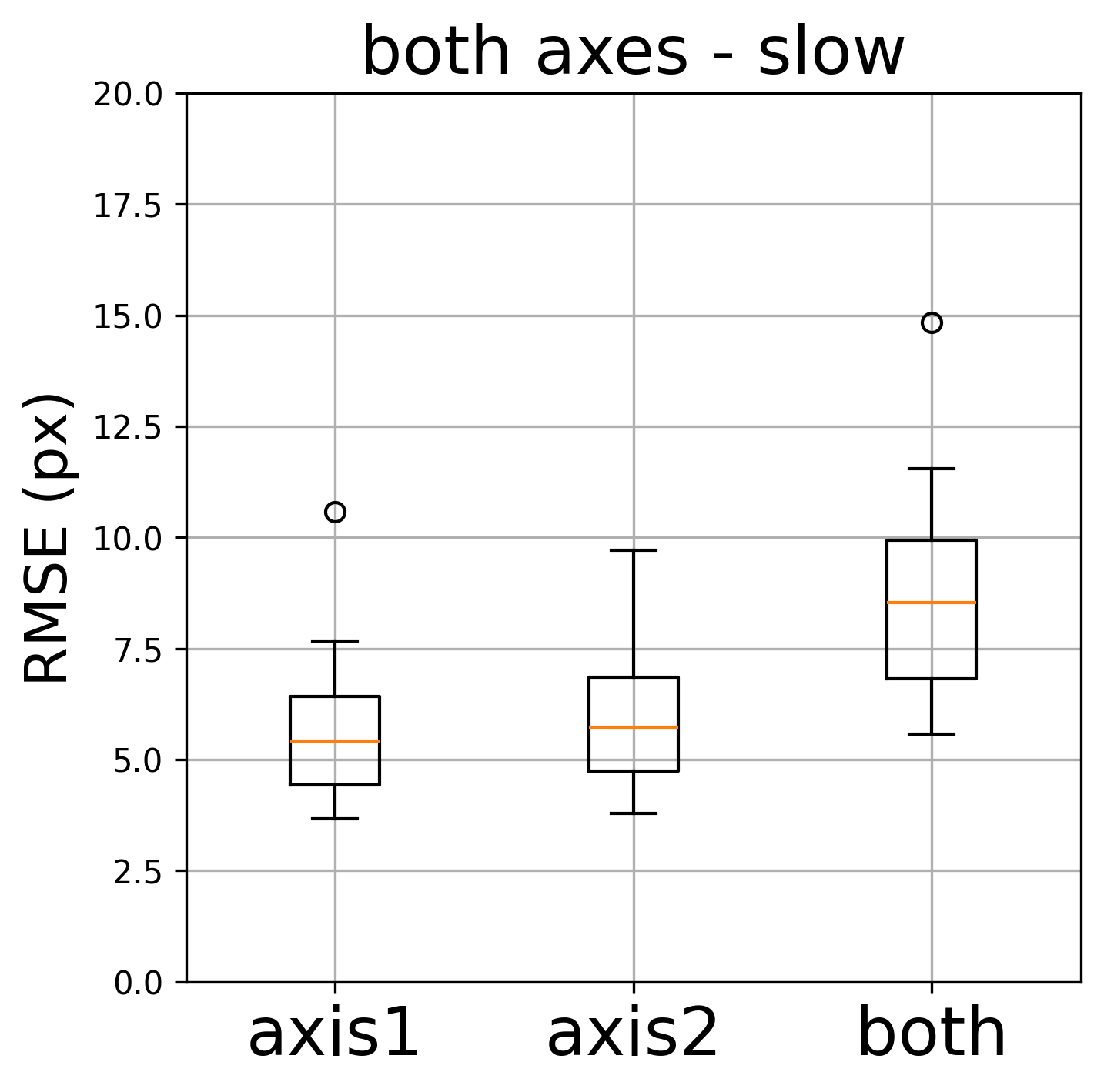} % Image 1, width relative to text width
        \hfill % Add horizontal space between images
        \includegraphics[width=0.32\textwidth]{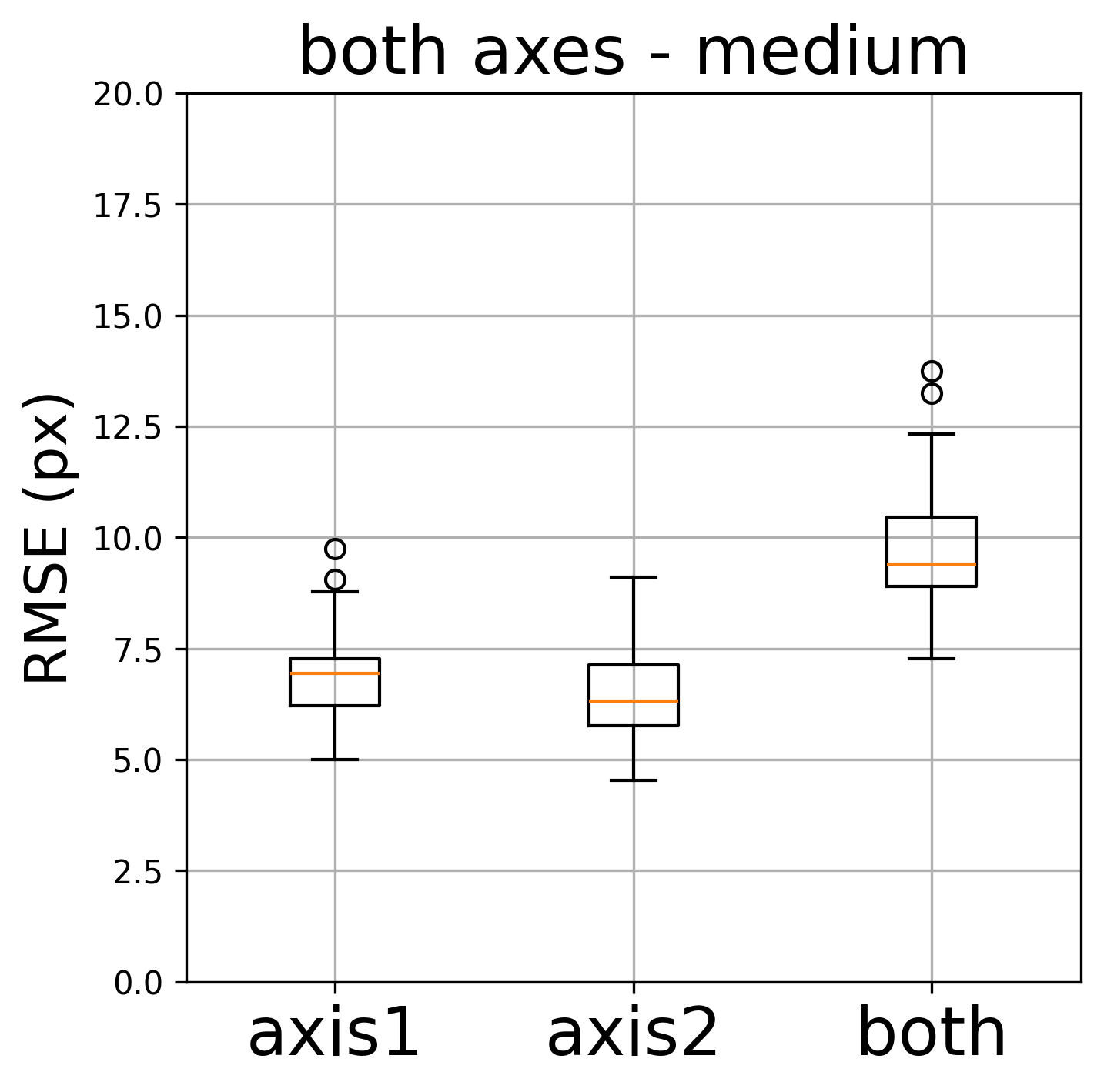} % Image 2, width relative to text width
        \hfill % Add horizontal space between images
        \includegraphics[width=0.32\textwidth]{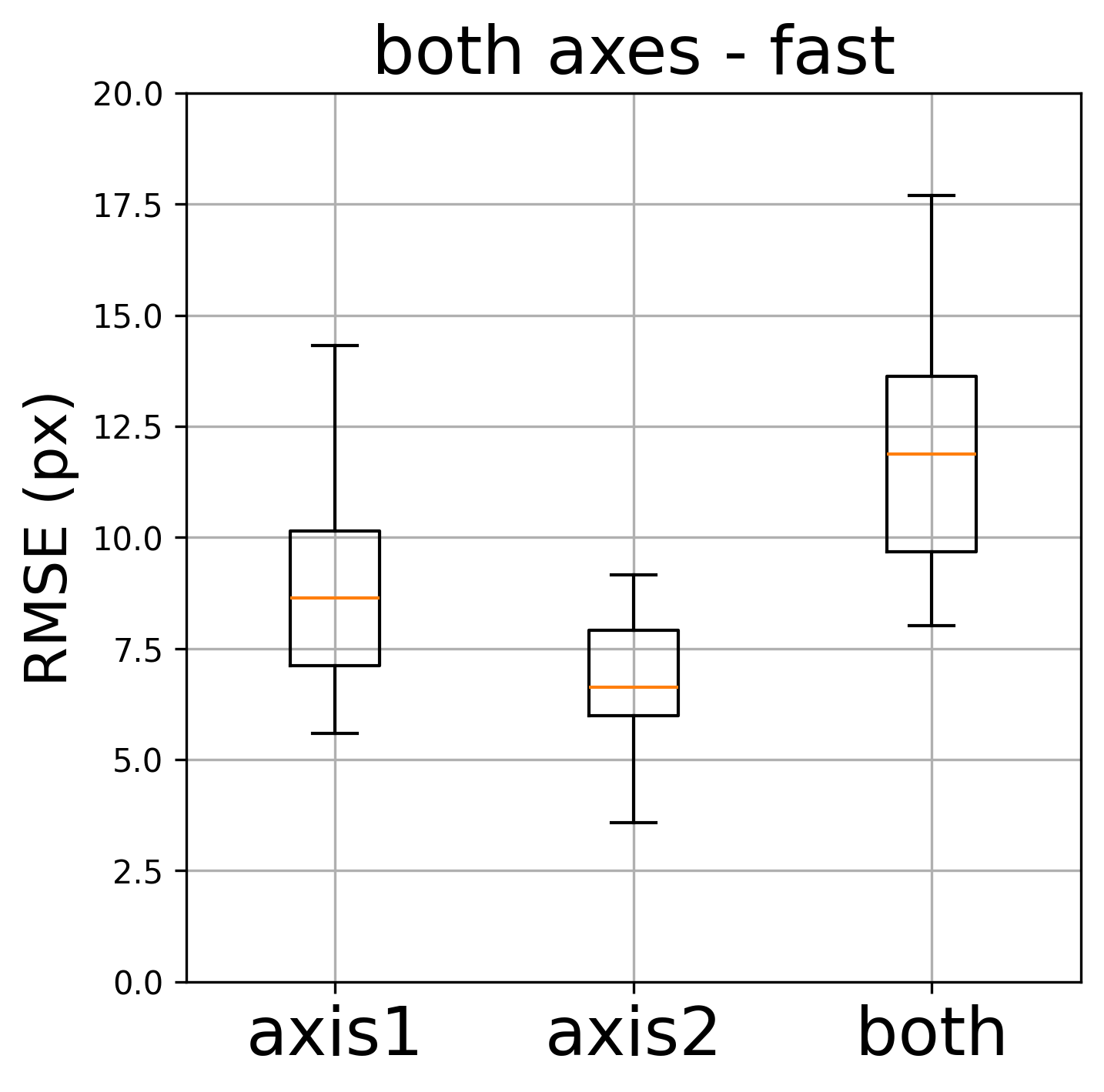} % Image 3, width relative to text width
        \caption{Analysis of jitter recovery for motion along \texttt{both axes}} % Caption for the entire third row
        \label{fig:row3}
    \end{subfigure}

    \caption{Performance of the propose algorithm over various frequencies and axial motion configuration.} % Caption for the entire figure
    \label{fig:allimages} % Label for the entire figure
\end{figure*}

\subsection{Performance Evaluation}
For each sequence in the dataset, we run the algorithm outlined in Sec.~\ref{sec:algorithm} to generate a jitter hypothesis at $t_{batch}$ intervals. However, due to the dynamics of the data collection process via the piezo controller, the ground truth is only recorded at 30 Hz, i.e., 33.3 ms intervals. The evaluation of algorithm's efficacy at recovering the current jitter needs to be evaluated at the corresponding timestamps of the ground truth. To bridge this gap and facilitate a meaningful comparison, we aggregate the jitter estimates generated within each ground truth interval. Specifically, we sum up the jitter estimates from the smaller $t_{batch}$ intervals to compute the total jitter over each $\delta t_{gt}$ period, where $\delta t_{gt}$ represents the time between consecutive ground truth measurements. This aggregation allows us to align our estimates with the temporal resolution of the ground truth data. Subsequently, we compare these accumulated jitter estimates against the corresponding ground truth values, enabling us to quantify the error between the estimated and actual jitter. This approach ensures a fair and accurate evaluation of our algorithm's performance, accounting for the different temporal resolutions of our estimates and the ground truth data.
% \begin{figure*}
%     \centering
%     \includegraphics[width=0.32\linewidth]{jitter-axis-first.png}
%     \includegraphics[width=0.32\linewidth]{jitter-axis-second.png}
%     \includegraphics[width=0.32\linewidth]{jitter-axis-both.png}
%     \caption{Error in estimated Jitter by speed: Analysis of sequences along each axis. \texttt{slow} sequence along each of the axis configurations has the least error. Fast sequences generally have higher errors along each axis of motion.}
%     \label{fig:statistics-results-speed}
% \end{figure*}

% \begin{figure*}
%     \centering
%     \includegraphics[width=0.32\linewidth]{jitter-speed-slow.png}
%     \includegraphics[width=0.32\linewidth]{jitter-speed-medium.png}
%     \includegraphics[width=0.32\linewidth]{jitter-speed-fast.png}
%     \caption{Error in estimated Jitter by the axis of motion: Analysis of sequences at different speeds. \texttt{second} axis has the least error most likely due to the differences between the underlying hardware.}
%     \label{fig:statistics-results-axis}
% \end{figure*}

\subsection{Recovered Jitter estimates}

\begin{figure*}
    \centering
    \includegraphics[width=0.32\textwidth]{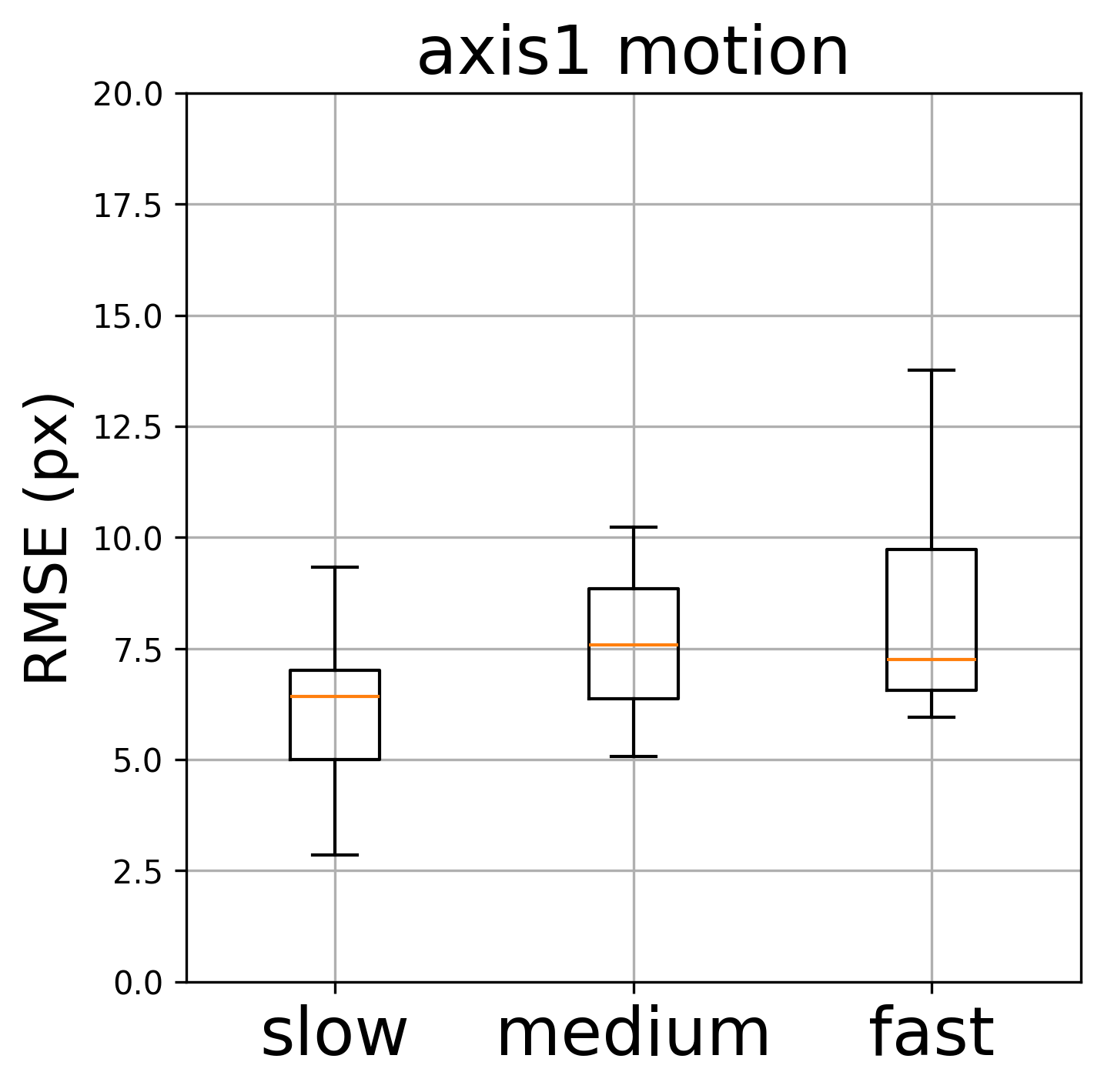} % Image 1, width relative to text width
    \hfill % Add horizontal space between images
    \includegraphics[width=0.32\textwidth]{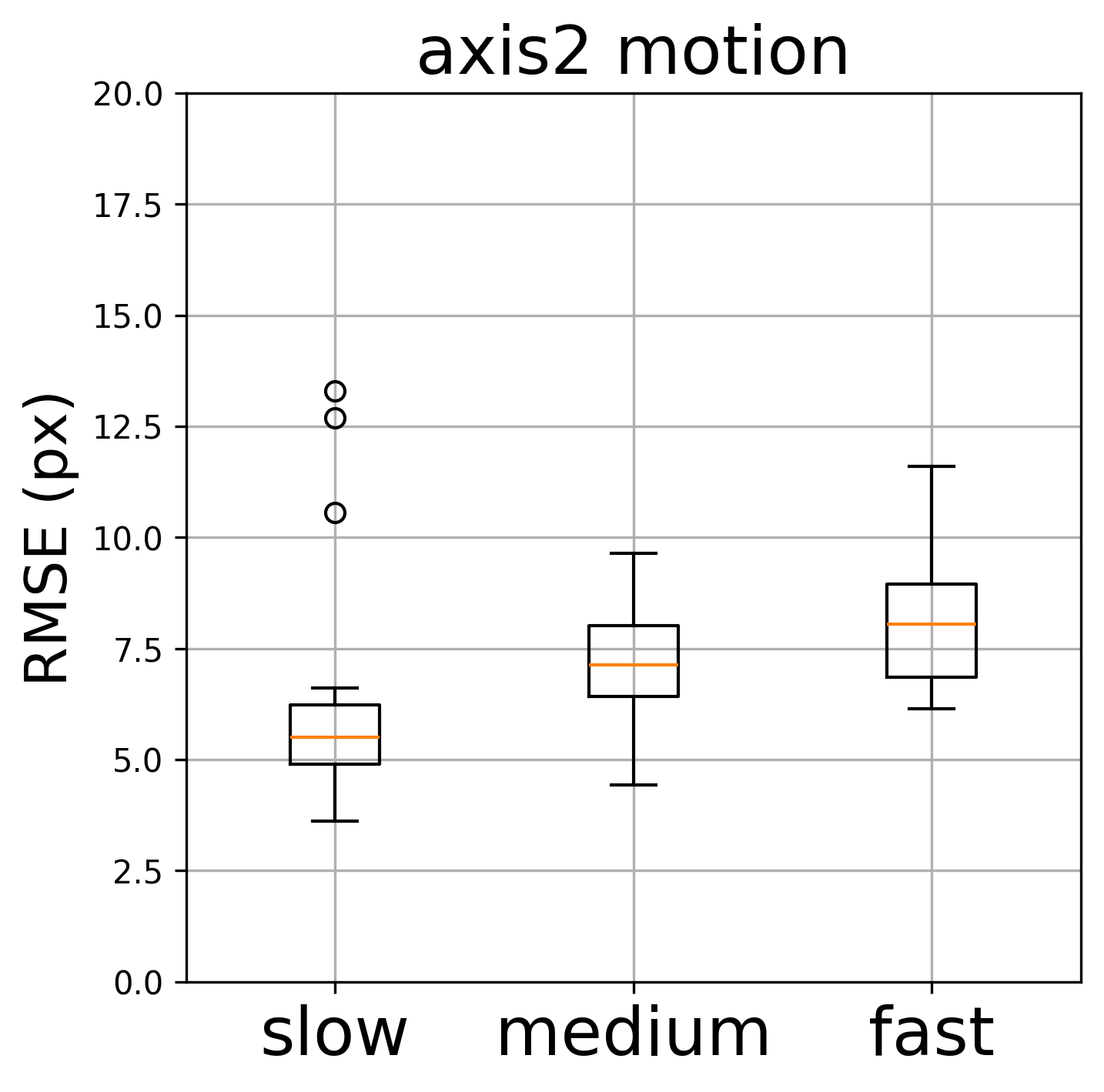} % Image 2, width relative to text width
    \hfill % Add horizontal space between images
    \includegraphics[width=0.32\textwidth]{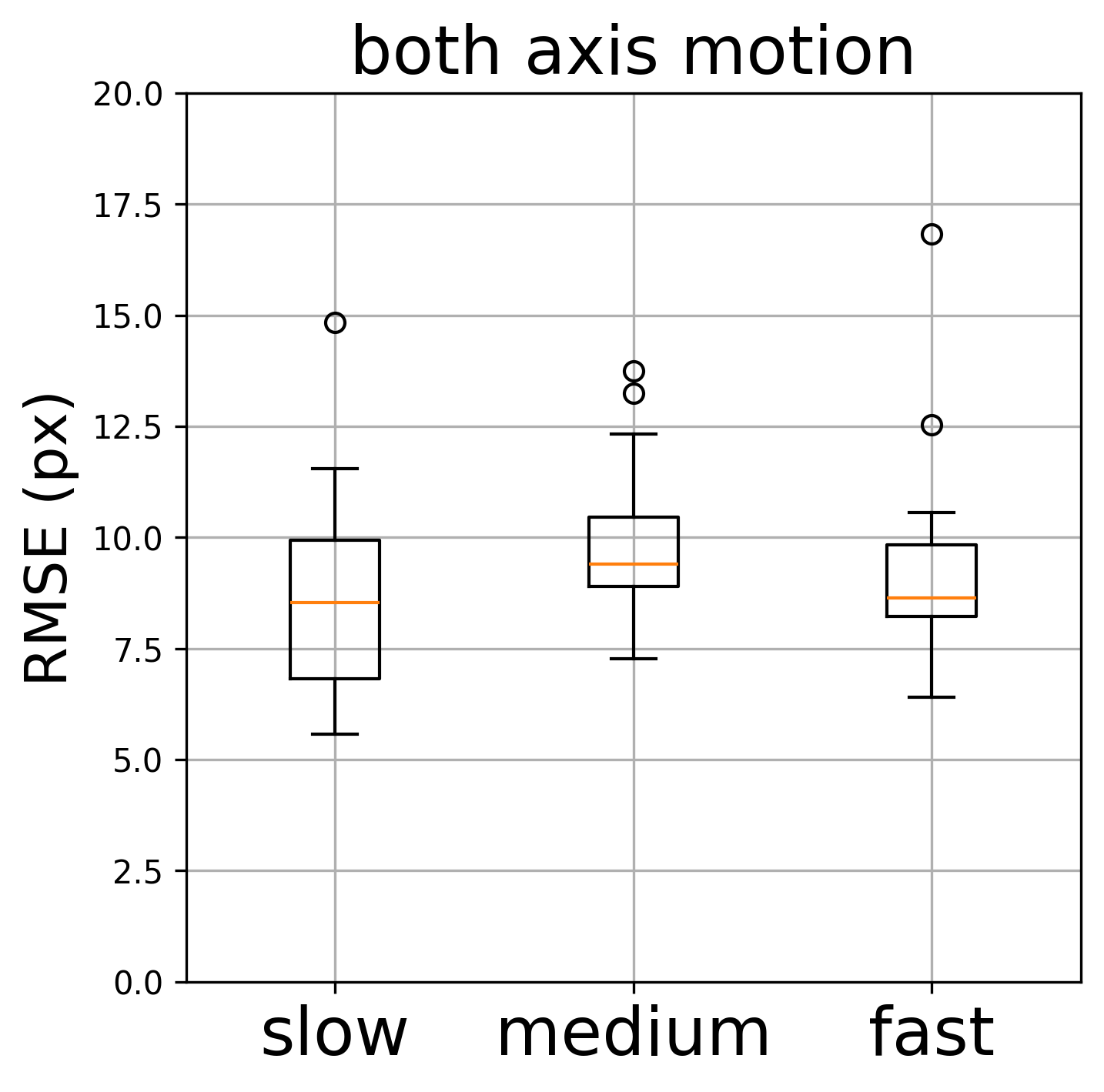} % Image 3, width relative to text width
    \caption{Summary statistics for jitter recovery for various motion profiles over all frequency settings.} % Caption for the entire 
    \label{fig:error_speed}
\end{figure*}
\begin{table*}
    \centering
    \begin{tabular}{cccc|ccc|ccc}
        & \multicolumn{9}{c}{Motion Axis and speeds} \\
        \cline{2-10}
        & \multicolumn{3}{c|}{Axis1} & 
        \multicolumn{3}{c|}{Axis2} & 
        \multicolumn{3}{c}{Both} \\
        % & \multicolumn{3}{c|}{speed} & \multicolumn{3}{c|}{speed} & \multicolumn{3}{c}{speed} \\
        \cline{2-10}
         & S & M & F & S & M & F & S & M & F  \\\hline\hline
        \multicolumn{1}{l|}{Error Axis1} & 6.14&7.67&8.19&3.57&2.82&1.48 & 7.82&8.88&8.76\\\hline
        \multicolumn{1}{l|}{Error Axis2}  &4.02&2.07&3.30&6.36&7.17&8.08&8.16&7.78&9.40\\\hline
        \multicolumn{1}{l|}{Error combined} &5.61&7.01&8.73&6.03&6.39&7.46&8.76&10.01&12.36\\\hline
        
    \end{tabular}
    \caption{Results for jitter recovery for the  e-STURT dataset. Errors along each individual axis as well as combined error is reported for all combination of motions and speeds.}
    \label{tab:results_main}
\end{table*}
\begin{figure*}
    \centering
    \includegraphics[width=0.32\linewidth]{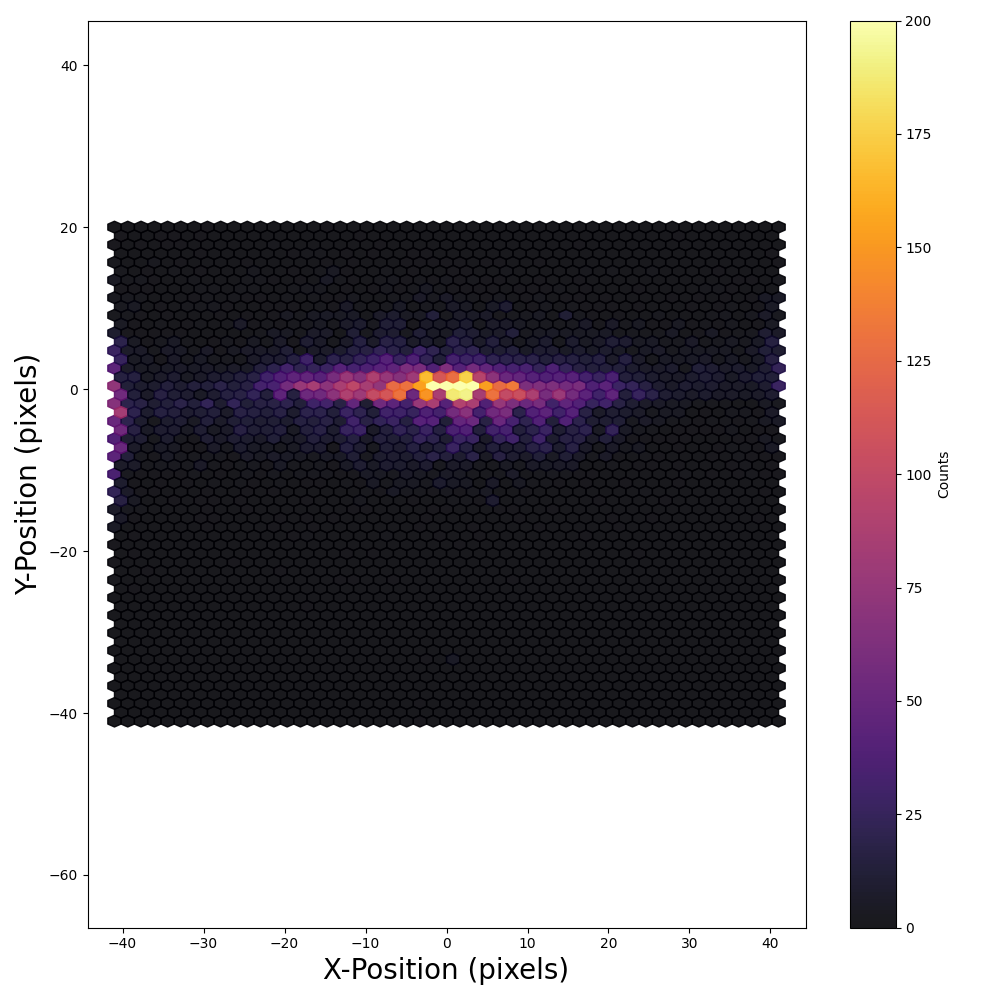}
    \includegraphics[width=0.32\linewidth]{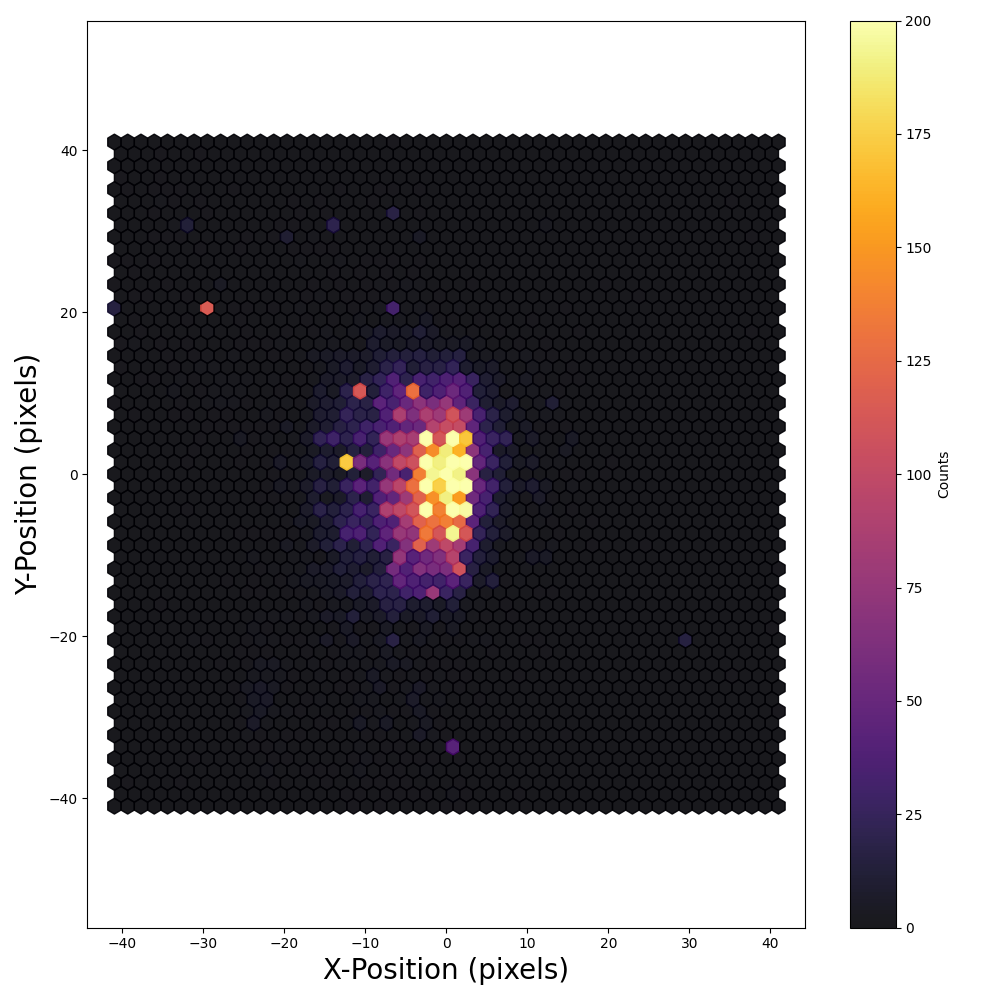}
    \includegraphics[width=0.32\linewidth]{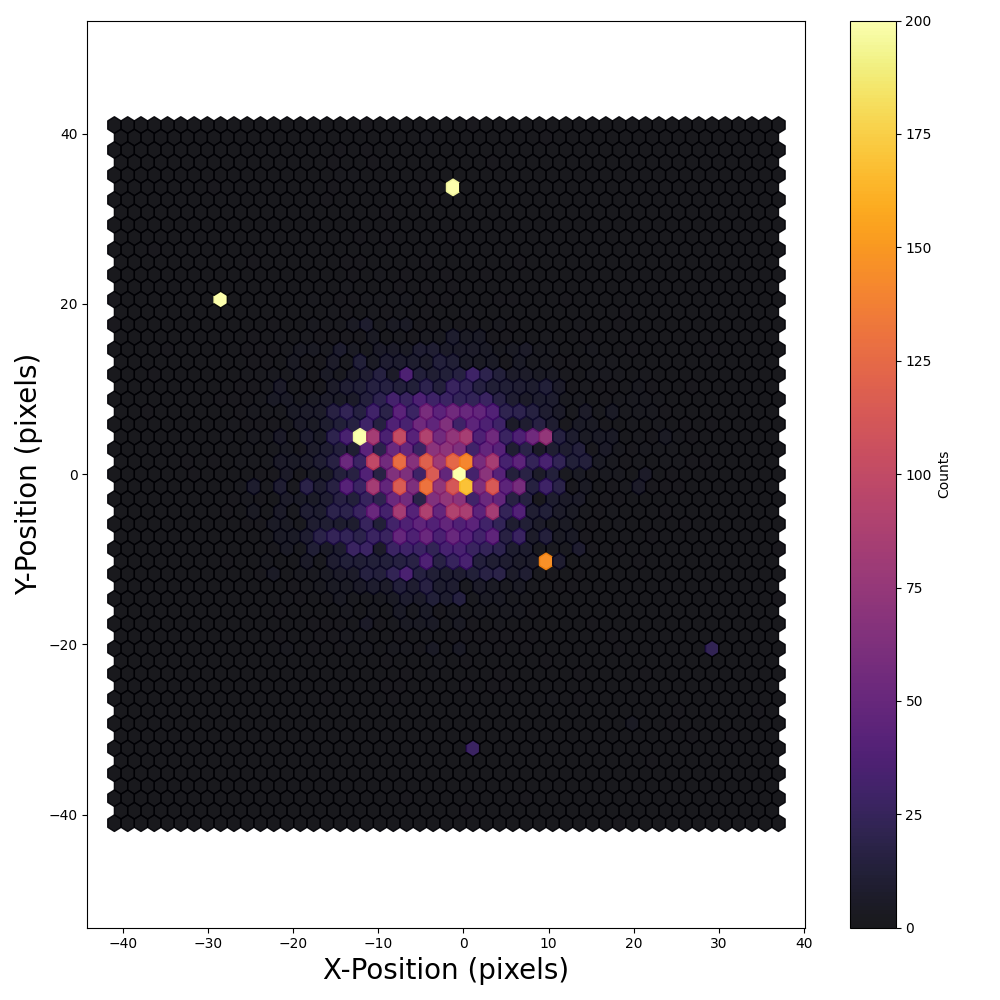}
    \caption{Heat maps of jitter estimates:\texttt{Axis1}, \texttt{Axis2}, and \texttt{BothAxes}. (truncated to show hypothesis with lower counts).}
    \label{fig:heatmaps}
\end{figure*}
We first depict the ground truth jitter and its estimate using the proposed algorithm in  Fig.~\ref{fig:x_y_plot}. It can be seen that, while noisy, the recovered estimate tracks the original underlying signal along both the axes. The frequency (speed) affects the fidelity with which the algorithm is able to recover jitter. For each of the speed settings, namely \texttt{slow}, \texttt{medium}, and \texttt{fast}, Fig.~\ref{fig:allimages} and  Table.~\ref{tab:results_main} report summary statistics for the performance of the algorithm.  It can be seen that for each axis setting, the algorithm is able to better recover jitter in the \texttt{slow} sequences. As more events are generated at the \texttt{slow} configurations, jitter estimation is relatively easier. At the higher speeds, performance degrades as the events required for jitter recovery become sparse, making it difficult to distinguish noise from signal. This can be seen in  Fig.~\ref{fig:error_speed} where results are summarized by axis of motion for different speeds.

The distribution of jitter amplitudes recovered by the algorithm is depicted in Fig.~\ref{fig:heatmaps} consisting of heatmaps for motion along the \texttt{Axis1}, \texttt{Axis2} and \texttt{Both Axes}. The amplitudes are distributed across the range of the observations (pixels). For both axes cases, the jitter amplitudes from a circle with the radius are defined by the maximum amplitude (0.1 mm / 20.58 pixels), demonstrating the algorithms ability to recover the whole range of motion inducted by jitter.

% In Fig.~\ref{fig:statistics-results-axis}, we analyze if the errors are symmetric along the different axes of the motion. It can be seen that errors at various speeds exhibit different characteristics, mostly due to the difference in hardware actuation by the piezoelectric stage. Errors are generally smaller at all speeds for the \texttt{second} axis, whereas the error for the \texttt{first} and \texttt{both axes} increase with the increase in the speed of motion.

\section{Conclusion and future work}
In this work, we present a comprehensive dataset consisting of event-camera sequences recorded under controlled jitter at various frequencies. We describe the hardware, the data generation method, and a baseline algorithm for jitter recovery that addresses the problems of noise filtering and motion recovery. The dataset will be made publicly available.

\bibliographystyle{IEEEtran}
\bibliography{citations}

\end{document}